\documentclass[10pt,journal,compsoc]{IEEEtran}

\ifCLASSOPTIONcompsoc
  \usepackage[nocompress]{cite}
\else
  \usepackage{cite}
\fi
\usepackage{multirow}
\usepackage{subfigure} 
\usepackage{graphicx}
\usepackage[safe]{tipa}
\usepackage{xcolor}
\usepackage[normalem]{ulem}

\ifCLASSINFOpdf
\else
\fi
\usepackage{amsmath}
\usepackage{amssymb}
\begin{document}

\title{Memory Augmented Deep Generative models for Forecasting the Next Shot Location in Tennis}
%
%
%
%

\author{Tharindu~Fernando,~\IEEEmembership{Student Member,~IEEE,}
        Simon~Denman,~\IEEEmembership{Member,~IEEE,}
       Sridha~Sridharan,~\IEEEmembership{Life Senior Member,~IEEE,}
         ~and~Clinton~Fookes,~\IEEEmembership{Senior Member,~IEEE}
         \IEEEcompsocitemizethanks{\IEEEcompsocthanksitem T. Fernando, S.Denman, S. Sridharan, C. Fookes  are with Image and Video Research Laboratory, SAIVT, Queensland University of Technology, Australia.\protect\\
E-mail: t.warnakulasuriya@qut.edu.au
}
\thanks{Manuscript received }}

%
%

\markboth{IEEE Transactions on Knowledge and Data Engineering}%
{Fernando \MakeLowercase{\textit{et al.}}: Bare Demo of IEEEtran.cls for Computer Society Journals}
%



\IEEEtitleabstractindextext{%
\begin{abstract}
This paper presents a novel framework for predicting shot location and type in tennis. Inspired by recent neuroscience discoveries we incorporate neural memory modules to model the episodic and semantic memory components of a tennis player. We propose a Semi Supervised Generative Adversarial Network architecture that couples these memory models with the automatic feature learning power of deep neural networks, and demonstrate methodologies for learning player level behavioural patterns with the proposed framework. We evaluate the effectiveness of the proposed model on tennis tracking data from the 2012 Australian Tennis open and exhibit applications of the proposed method in discovering how players adapt their style depending on the match context. 
\end{abstract}

}

\maketitle

\IEEEdisplaynontitleabstractindextext

%
\IEEEpeerreviewmaketitle

\IEEEraisesectionheading{\section{Introduction}\label{sec:introduction}}

%
%
%
%
\IEEEPARstart{T}{he} ability of professional athletes to accurately anticipate opponents' actions during a fast-ball sports such as tennis is considered a remarkable feat \cite{cacioppo2014intention}. Considering the fact that present day ball speeds exceed 130mph, the time required by the receiver to make a decision regarding the opponents' intention, and initiate a response could exceed the flight time for the ball \cite{crognier2005effect,cacioppo2014intention,williams2004developing,wright2007brain}.

Several studies have shown that this reactive ability is the product of pattern recognition skills that are obtained through a ``biological probabilistic engine'', that derives theories regarding opponents intentions with the partial information available \cite{cacioppo2014intention,wei2016forecasting,shvorin2017understanding}. For instance, it has been shown that expert tennis players are better at detecting events in advance \cite{williams2002anticipation,cacioppo2014intention} and posses better knowledge/ expertise of situational probabilities \cite{williams2004developing}. Further investigation of human neurological structures have revealed that those capabilities occur due to a bottom-up computational process \cite{cacioppo2014intention} within the human brain, from sensory memory to the experiences stored in episodic memory \cite{etzel2016brain,horzyk2017integration} and knowledge derived in semantic memory \cite{wang2017semantic,horzyk2017integration}.

Despite the growing interest among researchers in the machine learning domain in better understanding factors influencing decision making in fast-ball sports, there have been very few studies transferring the observations of the underlying neural mechanisms to neural modelling in machine learning. Current state-of-the-art methodologies try to capture the underlying semantics through a handful of handcrafted features, without paying attention to essential mechanisms in the human brain, where the expertise and observations are stored and knowledge is derived. It is a broadly established fact that handcrafted features only capture abstract level semantics in a given environment \cite{denton2016semi,ho2016generative,isola2017image,pan2017salgan} and it is proven that these ill represent the context in several data mining and knowledge discovery tasks \cite{fernando2015deep}. 

The goal of this paper is to derive a deep learning model to anticipate the next shot location in tennis, given current match context and a short history of player and ball behaviour from Hawk-eye ball tracking data \cite{hawk_eye}. Our predictions comprise the next shot location as well as the type of the shot to anticipate.

Inspired by the automatic loss function learning power of Generative Adversarial Networks (GAN) \cite{ho2016generative,our_wacv1,arici2016associative,goodfellow2014generative,pan2017salgan,yoo2016pixel,zhao2016energy} and the capability of neural memory models \cite{our_wacv1,kumar2016ask,parisotto2017neural,wang2017semantic} to store and retrieve semantic level abstractions from historical agent behaviour, we propose a Memory augmented Semi Supervised Generative Adversarial Network (MSS-GAN). We demonstrate that the proposed framework can be utilised not only for high performance coaching \cite{whitmore2010coaching,reid2008quantification}, and designing intelligent camera systems for automatic broadcasting \cite{wang2004automatic,carr2013hybrid} where the system anticipate the next shot and shot type to better capture the player behaviour; but also for better understanding of player strategies, strengths and weaknesses. 

Due to the conditional nature of the proposed framework, the derived knowledge from the proposed memory modules can be utilised for demonstrating player behaviour changes when encountering different contexts. The main contributions of the proposed work are summarised as follows: 

\begin{enumerate}
\item We introduce a novel end-to-end deep learning method that learns to anticipate player behaviour.
\item We propose a Semi Supervised Generative Adversarial Network architecture that is coupled with neural memory modules to jointly learn to generate the return shot trajectory and to classify the shot type.
\item We demonstrate how the proposed framework could be utilised to infer player styles and opponent adaptation strategies. 
\item We perform an extensive evaluation of the proposed method on tennis player tracking data from the 2012 Australian Tennis open.
\item We provide comprehensive analysis on the contribution of each component in the proposed framework by evaluating the proposed method against a series of counterparts.   
\end{enumerate}
 
\section{Related Work}

\subsection{Sports Prediction}
With the recent advancements in ball and player tracking systems in sports, there has been increasing interest among researchers to utilise this data in numerous data mining and knowledge discovery tasks. 

In \cite{cervone2014pointwise} the authors utilise a possession value model for predicting points and behaviour of the ball handler in basket ball, assuming that the player behaviour depends only on the current spatial arrangement of the team. In \cite{carr2013hybrid} a future player location prediction strategy is applied to move a robotic camera, with applications to automatic broadcasting. Wei et al. \cite{wei2013predicting} proposed a graphical model for predicting the future shot location in tennis. They  utilise handcrafted, dominance features together with the ball bounce location, ball speed and player feet locations when determining the future player behaviour. This model is further augmented in \cite{wei2016forecasting} where the authors utilise a Dynamic Bayesian Network to model the same set of features. 

However recent studies in the sports prediction field \cite{fernando2015deep} have demonstrated the importance of learning the underlying feature distribution in an automatic fashion. For instance in \cite{fernando2015deep} the authors learn a dictionary of player formations in soccer for classifying the outcome of a shot. Even though they achieve comprehensive advancement towards automatic feature learning with player trajectories, those systems cannot be directly applied to model player strategies in tennis. When anticipating future player behaviour, based on the neurological observations presented in \cite{williams2002anticipation,cacioppo2014intention} it is vital to incorporate a player's past experiences and derived knowledge into the context modelling process. 

\subsection{Neural Memory Networks}
A memory module is required to store important facts from historical information. Neural memory modules have been extensively applied in numerous domains \cite{malinowski2014multi,our_wacv1,parisotto2017neural,fernando2017Learning}, where the model learns to automatically store and retrieve important information that is vital for the prediction task. 

In the reinforcement learning domain, Horzyk et al. \cite{horzyk2017integration} proposed an episodic memory architecture composed of a tree-structured memory. In a similar line of work, the authors in \cite{wang2017semantic,laird2014case,mueller2006rem,rinkus2004neural} investigate possible interaction structures for semantic memory. However all these frameworks are proposed in the reinforcement learning domain and a substantial amount of re-engineering is required to adapt those strategies to the supervised learning domain. Furthermore, adaptation of the structure is necessary for modelling player specific knowledge. 

We build upon the tree-memory structure proposed in \cite{fernando2017tree}. The authors in \cite{fernando2017tree} propose this model to map longterm temporal dependencies. We expand this memory structure and propose a novel neural memory structure for episodic memory (EM) and suggest a framework for multiple memory interactions and propagating the knowledge from the EM to the semantic memory (SM). 

\subsection{Generative Adversarial Networks}
Generative adversarial networks (GAN) belong to the family of generative models, and have achieved encouraging results for image-to-image synthesis \cite{arici2016associative,our_wacv1,isola2017image}. These models partake in a two player adversarial game where the Generator (G) ties to fool the Discriminator (D) with synthesised outputs while the D tries to identify them. 

There exist numerous architectural augmentations for GANs. For instance, in \cite{yu2017seqgan} the authors utilise a recurrent network approach for handling sequential data. Most recently authors in \cite{pan2017salgan} have utilised the GAN architecture for visual saliency prediction and further augmented it with memory architectures in \cite{our_wacv1} for capturing both low and high level semantics in modelling human gaze patterns. 

We are inspired by the Semi-Supervised conditional GAN (SS-GAN) proposed in \cite{denton2016semi}, where the authors couple the unsupervised loss of the GAN together with a supervised classification objective. The authors have shown this to enhance the generator's performance by incorporating class specific semantics into the synthesis process. We enhance the SS-GAN model by coupling it with neural memory networks by drawing parallels with recent neurological observations, and propose avenues to achieve player level adaptations of the model. 

\section{Architecture}
We are motivated by the neuroscience observations provided in \cite{cacioppo2014intention}. They present strong evidence towards activations of brain areas known to be involved with perception: Episodic Memory (EM) where personal experiences are used to determine the similarities between current sensory observation and the stored experiences \cite{etzel2016brain,tulving1985elements}; and Semantic Memory (SM) where foundations of knowledge and concepts are stored \cite{wang2017semantic,horzyk2017integration}. 

\begin{figure*}[htbp]
\centering
{\includegraphics[width = .95 \linewidth]{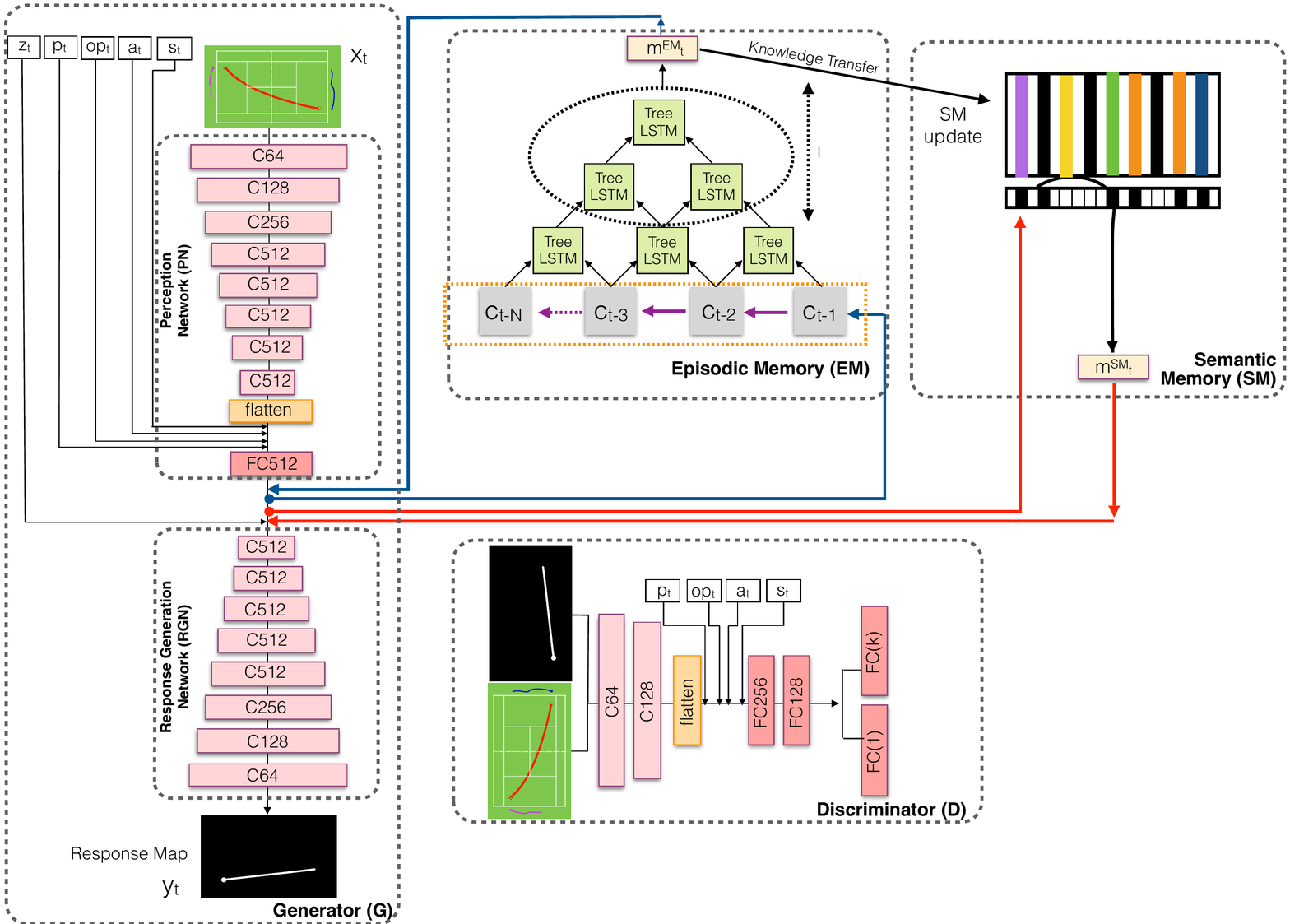}}
\caption{Proposed MSS-GAN model: The model is composed of the \textbf{Perception Network (PN)}  which encodes the visual input and concatenates it together with sparse speed ($s_t$), angle ($a_t$), opponent id ($op_t$) and point ($p_t$) representations; \textbf{Episodic Memory (EM)} which stores the temporally adjacent player experience embeddings; \textbf{Semantic Memory (SM)} which extracts out knowledge from the EM and a \textbf{Response Generation Network (RGN)} which generates a future shot location based on these observations with the aid of a latent noise distribution $z_t$. The \textbf{Discriminator (D)} receives the input perception together with the generated response from the RGN and determines whether the response is a true player behaviour or a synthasised one from the RGN, and classifies the shot type. }
\label{fig::fig_model}
\end{figure*}

Figure \ref{fig::fig_model} shows the overall structure of the proposed approach. The Perception Network (PN) (see Sec. \ref{sec:pn}) processes incoming images to obtain an embedding that represents the shot. This is combined with embeddings from the Episodic Memory (EM) (Sec. \ref{sec:em}) and Semantic Memory (SM) (Sec. \ref{sec:sm}) to predict the next shot via the Response Generation Network (RGN) (Sec. \ref{sec:rgn}). Note that the output of the PN is also fed to the memories to learn historic behaviour. Finally, we use a GAN framework to learn the network, with the predicted shot from the RGN being passed to the Discriminator to determine if it is a realistic shot or not.  

For the rest of the paper we use the following notation. All weights are denoted $W$, $f$ denotes forget gates and $F^{NE}(X)$ denotes a function that passes input $X$ through one of more layers of the network $NE$. $LSTM$ denotes a layer consisting of LSTM cells. $c_t$ represents the current state (context) of the game. The current representation of the memory at time instant $t$ is denoted by $M_{t}$, while the query vector that is used to read the memory is denoted by $q_t$ and $m_t$ represents the output of the memory read operation. Attention values, $\eta$, denote the attention given to the content of $M_t$ to answer the query vector $q_t$, while the normalised attention values are denoted by $\alpha$.

\subsection{Perception Network (PN)}
\label{sec:pn}
Our observations are from Hawk-eye player tracking data \cite{hawk_eye}, which stores the ball trajectory and player feet movements along with the ball speed and angle. For each shot event in the database we extract out the ball and player trajectory from the shot start time to the present, and generate an image depicting the perception of the shot receiver. 

\begin{figure}[htbp]
\centering
\subfigure[]{\includegraphics[width = .35 \linewidth]{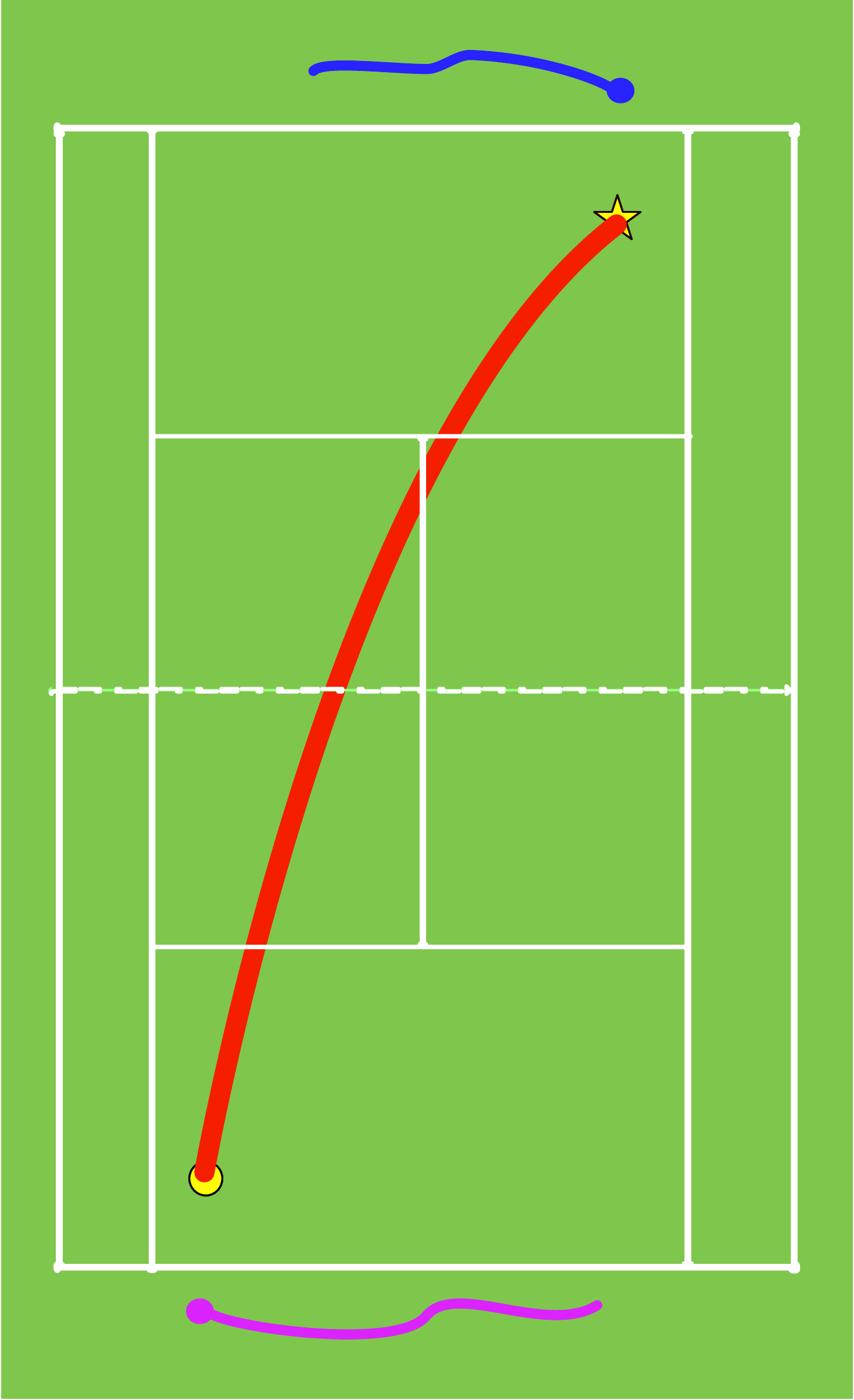}}
\subfigure[]{\includegraphics[width = .35 \linewidth]{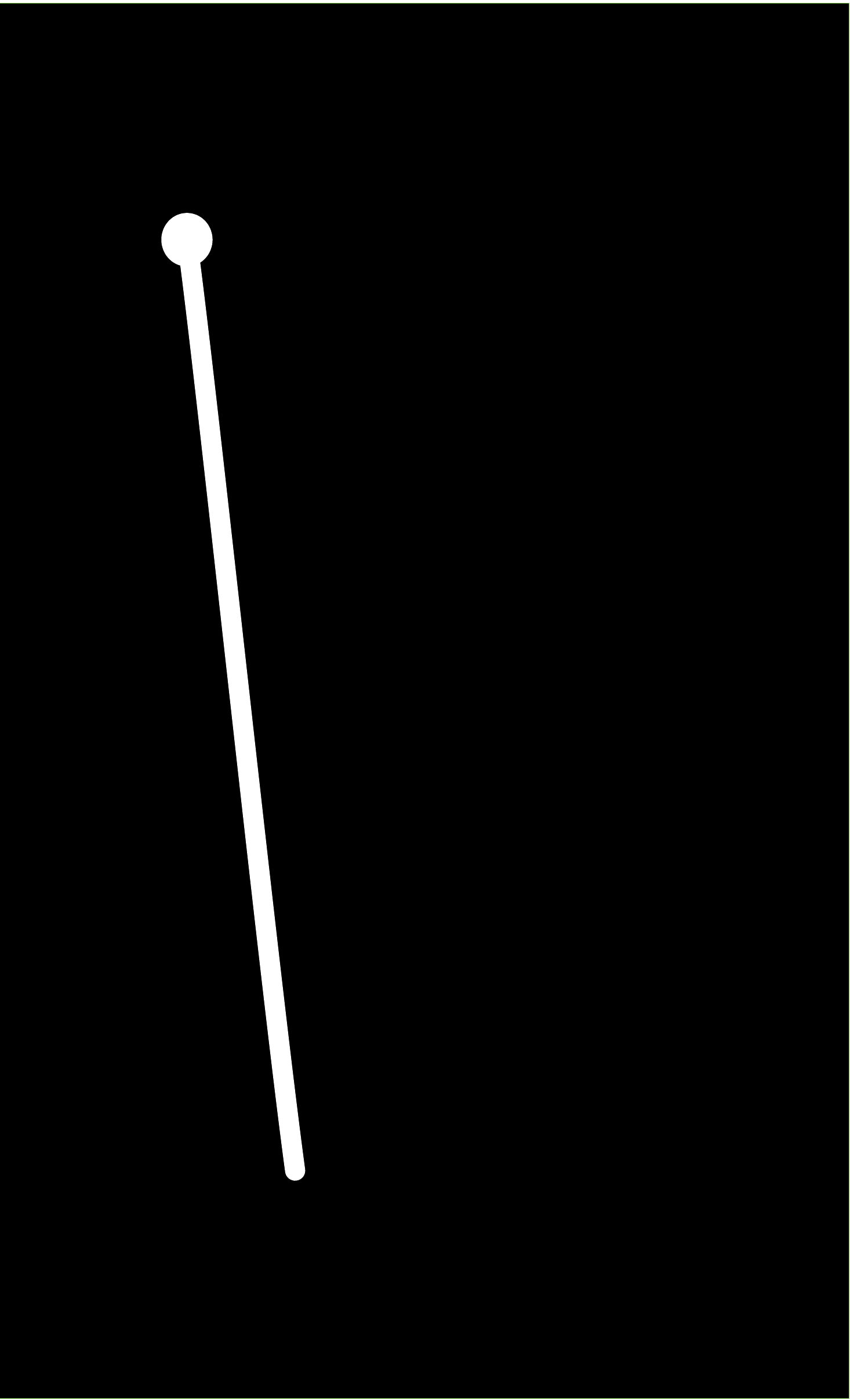}}
\caption{A sample input and output for the proposed MSS-GAN model. The observed incoming ball trajectory is denoted in red where the starting and ending locations are presented with yellow star and a circle. The opponent and player feet movements are denoted in blue and magenta colours where the circle denotes the ending location of the respective player trajectory. The predicted next shot is shown in (b) and is denoted with a white line where the ball landing location is given by a white circle.}
\label{fig:fig_input_output}
\end{figure}

Fig. \ref{fig:fig_input_output} (a) illustrates the input to our perception network. The opponent trajectory is denoted in blue where the circle denotes the ending location, and the ball trajectory is shown in red. Shot starting and ending locations are denoted with a  yellow star and a circle, respectively. In order to account for the current position of the shot receiver we encode his trajectory in magenta where the circle denotes the ending location. A sample output generated by the proposed framework is given in Fig. \ref{fig:fig_input_output} (b), where the ball trajectory for the predicted next shot is denoted with a white line. We utilise images to represent our observations as they preserve the relative spatial relationships \footnote{Please note that in Fig. \ref{fig:fig_input_output} (b) we have used black and white masks without court outlines as it reduces the number of parameters in the prediction module that requires training.}.

Fig. \ref{fig::fig_model} shows the architecture of the proposed Perception Network (PN). Deep Convolution Neural Networks (DCNN) have shown encouraging results when encoding information via automatic feature learning. Hence we utilise a C64-C128-C256-C512-C512-C512-C512-C512 DCNN structure in our PN where  Ck denote a Convolution-BatchNorm-ReLU layer with k filters. We then pass this embedding through a fully connected layer with 512 neurons, (FC512), which concatenates the image representation with the current incident speed, $s_t$, angle, $a_t$, opponent id, $op_t$, and the current player points, $p_t$, in order to generate the current state embedding $c_t$. 

\subsection{Episodic Memory (EM)}
\label{sec:em}
When considering the human cognitive structure, EM is vital for storing spatio-temporal event information, for helping to form concepts in Semantic Memory (SM), and for guiding the response generation \cite{rinkus2017superposed}. EM is composed of one's accumulated past experiences, and contains the encoded event information such as what, where and when. This allows us to mentally re-visit past experiences and generate observations comparing them with the current state and respond accordingly \cite{horzyk2017integration}. 

Hence, drawing parallels to a tennis tournament, we pass the observations of each player separately through the PN and store the embedded observations in a memory queue. 

When deriving long term relationships among the stored sequential data, Fernando et. al \cite{fernando2017tree} have shown tree-LSTM cells to preserve adjacent temporal relationships and propagate salient information effectively, aiding decision making in the present state. Hence we utilise tree-LSTM cells to summarise the content of our EM queue, arranging the content in a tree structure. 

The functionality of the EM module can be summarised by the operations Memory Read (see Section 3.2.1) and Memory Write (3.2.2). 

\subsubsection{Memory read}
Let $\ddot{x}_t=[x_t, s_t, a_t, op_t, p_t]$, where $x_t$ is the current observation image, $s_t$ is the incident speed, $a_t$ is the incident angle, $op_t$ is the opponent id and $p_t$ is the points for shot receiver and the opponent. Then the encoded vector for the current state representation from the PN network at time step $t$ can be denoted as,

\begin{equation}
c_t = F^{PN}(\ddot{x}_t),
\end{equation}

where $F^{PN}$ is the Perception Network (see Sec. \ref{sec:pn}) and  $c_t  \in  \mathbb{R}^{1 \times k}$. Consider the EM to have $N$ embeddings stored, with each having the size $1 \times K$. Similar to \cite{fernando2017tree}, when computing the EM output at time instance $t$, we extract the memory tree configuration at time instance $t-1$. Let $M_{(t-1)}^{EM} \in \mathbb{R}^{k \times 2^{l}}$ be the memory matrix resultant from concatenating nodes from the top of the tree to depth $l = [1, \ldots]$. 

The tree memory architecture hierarchically maps the memory with a bottom up tree structure. The bottom layer of the tree stores all historic states and the hierarchy is created such that the most significant features are concatenated, propagating two temporally adjacent neighbours to the upper layer. This can be seen as each layer performing information compression. Hence the top most layer contains the most compressed version of the information present in the memory. In the immediately preceding layer this compression is slightly relaxed.  Using the depth ($l$) hyper-parameter we extract out information from multiple levels from the tree top, allowing us to extract different levels of abstraction. $l$ ranges from $l=1$ to the total number of layers in the tree hierarchy, where 1 denotes that the information is extracted only from the tree top.  The optimal value for $l$ is evaluated experimentally and this evaluation is presented in Fig. \ref{fig:hyper_parameters} (b).

The read head on the EM passes $c_t$ through a read LSTM function, $LSTM^{EM,r}$, to generate a vector to query the memory such that,
\begin{equation}
q_t^{EM} = LSTM^{EM,r}(c_t,M^{EM}_{(t-1)}),
\end{equation}
Then an attention vector $\eta_t^{EM}$ determines the similarity between the current context vector $c_t$ and the memory representation $M_{(t-1)}^{EM}$ by attending over each element such that,
 \begin{equation}
\eta_{(t,j)}^{EM} =q_t^{EM}M^{EM}_{(t-1,j)},
\end{equation}
where $M^{EM}_{(t-1,j)}$ denotes the $j^{th}$ item in the matrix $M^{EM}_{(t-1)}$, $j= [1, \ldots, 2^{l}-1]$. Then the score values are normalised using soft attention \cite{duan2017one,xu2015show,fernando2017Learning}, generating a probability distribution over each memory element as follows,
\begin{equation}
 \alpha_{(t,j)}^{EM}=\dfrac{\mathbb{E}(\eta_{(t, j)}^{EM})}{ \sum_{j=1}^{2^l-1} \mathbb{E}(\eta_{(t, j)}^{EM})}.
\label{eq:h_t}
\end{equation}
Now we generate the output of the memory read by,
\begin{equation}
 m_t^{EM}=\sum_{j=1}^{2^l-1} \alpha^{EM}_{(t,j)}M_{(t-1, j)}^{EM}.
\label{eq:mem_out}
\end{equation}

\subsubsection{ Memory Write}
\begin{figure}[t]
\center
\includegraphics[width = .7\linewidth]{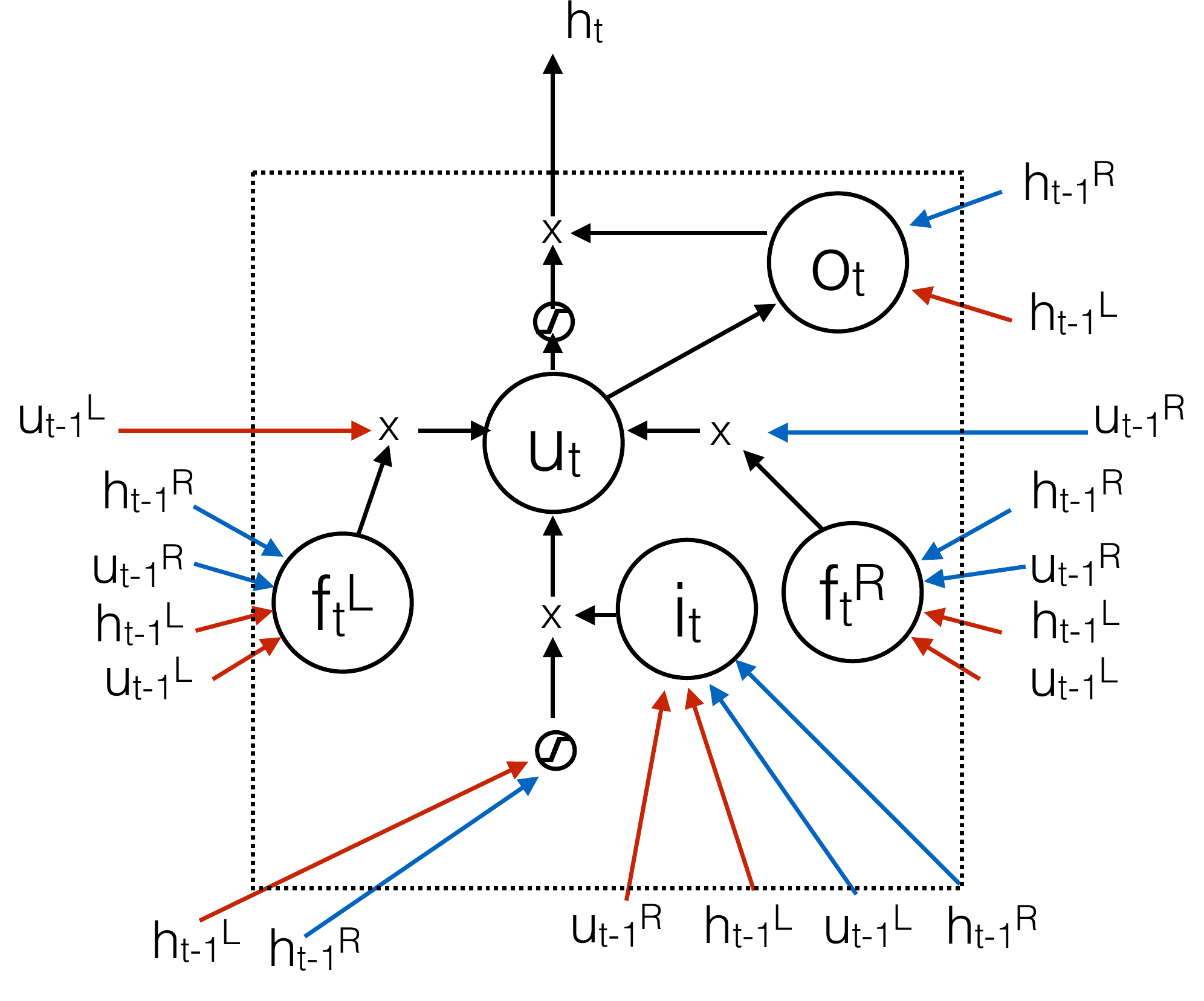}
\caption{Tree memory cell architecture. $f_t^L, f_t^R, o_t, i_t$ represents the left forget gate, right forget gate, output gate and input gate respectively. $\times$ represents multiplication}
\label{fig:memory_cell}
\end{figure}

The new encoded information, $c_t$, is appended to the end of the EM queue. This invokes the memory update operation which updates the content of the tree-LSTM cells. Each memory cell contains one input gate, $i_t$, one output gate, $o_t$, and two forget gates $f_t^L$ and $f_t^R$ for left and right child nodes. At time instance $t$ each node in the memory network is updated in the following manner, 

 \begin{equation}
i_t=\sigma(W_{hi}^Lh_{t-1}^L + W_{hi}^Rh_{t-1}^R +W_{ui}^Lu_{t-1}^L +W_{ui}^Ru_{t-1}^R ) ,
\label{eq:i_t}
\end{equation}
\begin{equation}
f_t^L=\sigma(W_{hf_l}^Lh_{t-1}^L + W_{hf_l}^Rh_{t-1}^R +W_{uf_l}^Lu_{t-1}^L +W_{uf_l}^Ru_{t-1}^R ) ,
\end{equation}
\begin{equation}
f_t^R=\sigma(W_{hf_r}^Lh_{t-1}^L + W_{hf_r}^Rh_{t-1}^R +W_{uf_r}^Lu_{t-1}^L +W_{uf_r}^Ru_{t-1}^R ) ,
\end{equation}
\begin{equation}
\beta=W_{hu}^Lh_{t-1}^L+ W_{hu}^Rh_{t-1}^R ,
\end{equation}
\begin{equation}
u_t^P=f_t^L \; \times \; u_{t-1}^L + f_t^R \; \times \;  u_{t-1}^R + \;  i_t  \; \times \; tanh(\beta) ,
\end{equation}
\begin{equation}
o_t=\sigma(W_{ho}^Lh_{t-1}^L + W_{ho}^Rh_{t-1}^R +W_{uo}^Pu_{t}^P) ,
\end{equation}
\begin{equation}
h_t^P=o_t \;  \times \; tanh(u_t^P) ,
\label{eq:h_t}
\end{equation}

where $h_{t-1}^L $,  $h_{t-1}^R $, $u_{t-1}^L$ and $u_{t-1}^R$ are the hidden vector representations and cell states of the left and right children respectively; and $h_t^P$ and $u_t^P$ are the hidden state and cell state representations of the parent node. The relevant weight vectors, $W$, are represented with appropriate super and subscripts where the superscript represents the relevant child node, and the subscript represents the relevant gate the weight is attached to. The process is illustrated in Fig \ref{fig:memory_cell}.

\subsection{Semantic Memory}
\label{sec:sm}
According to Wang et. al \cite{wang2017semantic} semantic memory can be considered as the derived knowledge from the specific experiences stored in episodic memory. It doesn't contain any situational information such as what, where and when. However it contains derived salient information from the sequential information. 

Hence, in contrast to EM, which is constructed as a queue representing the temporal structure, we design SM as matrix of $b$ elements $M_t^{SM} \in \mathbb{R}^{k \times b}$ where $k$ is the embedding dimension of $c_t$.

\subsubsection{Memory Read}
The read operation of the SM is identical to the read operation of the EM. Formally, let the content of the SM at time instance $t-1$ be $M_{t-1}^{SM}$. Then the memory read operation can be summarised as,

\begin{equation}
q_t^{SM} = LSTM^{SM,r}(c_t,M^{SM}_{(t-1)}),
\end{equation}
 \begin{equation}
\eta_{(t,j)}^{SM} =q_t^{SM}M^{SM}_{(t-1,j)},
\end{equation}
\begin{equation}
 \alpha_{(t,j)}^{SM}=\dfrac{\mathbb{E}(\eta_{(t, j)}^{SM})}{ \sum_{j=1}^{b} \mathbb{E}(\eta_{(t, j)}^{SM})}.
\label{eq:h_t}
\end{equation}
\begin{equation}
 m_t^{SM}=\sum_{j=1}^{b} \alpha^{SM}_{(t,j)}M_{(t-1, j)}^{SM},
\end{equation}
where $j = [1, \ldots, b]$.

\subsubsection{Memory Write}
The memory update procedure of the EM triggers the update procedure of the SM. Let the output of the EM tree at time instance $t$ be denoted by $m_t^{EM}$. We generate a vector $\grave{m}_{t}$ for the SM update by passing the output of the memory through a write LSTM function $LSTM^{SM,w}$, 

\begin{equation}
\grave{m}_{t} = LSTM^{SM,w}(m_t^{EM}),
\end{equation}

Then we generate the attention score values,
\begin{equation}
\grave{\eta}_{(t,j)}^{SM} =\grave{m}_t^{SM}M^{SM}_{(t-1,j)},
\end{equation}

and normalise them as,
\begin{equation}
 \grave{\alpha}_{(t,j)}^{SM}=\dfrac{\mathbb{E}(\grave{a}_{(t, j)}^{SM})}{ \sum_{j=1}^{b} \mathbb{E}(\grave{a}_{(t, j)}^{SM})}.
\label{eq:h_t}
\end{equation}
Then we update the SM as,
\begin{equation}
 M_t^{SM}=M_{t-1}^{SM}(I-\grave{\alpha}_t \otimes e_k)^T + (\grave{m}_t \otimes e_b)(\grave{\alpha}_t \otimes e_k)^T,
\label{eq:mem_write}
\end{equation}

where $I$ is a matrix of ones, $e_b \in \mathbb{R}^{b}$ and $e_k \in \mathbb{R}^{k}$ are vector of ones and $\otimes$ denotes the outer product which duplicates its left vector $b$ or $k$ times to form a matrix. 

\subsection{Response Generation Network (RGN)}
\label{sec:rgn}
The proposed RGN takes a latent noise distribution $z_t$ together with the embedded input vector $c_t$ and the memory output vectors $m_t^{EM}$ and $m_t^{SM}$, and generates the response map, $y_t$, denoting the ball trajectory of the next shot. Our RGN has the structure CD512-CD512-CD512-C512-C256-C128-C64 where CDk denotes a a Convolution-BatchNormDropout-ReLU layer with a dropout rate of 50\%.  The RGN generates a response map as illustrated by Fig. \ref{fig:fig_input_output} (b).

\section{Model Learning}

Most recently, Generative Adversarial Networks (GAN) have shown exemplary results in image to image synthesis problems \cite{isola2017image}. GANs are comprised of two components: a Generator,$G$, and a Discriminator, $D$, competing in a two player game. We draw our inspiration from the Semi-Supervised conditional GAN (SS-GAN) \cite{denton2016semi} architecture where $G$ receives the observed state representation $\ddot{x}_t$ and a random noise vector $z_t$ and tries to synthesise the response map $y_t$: $G(\ddot{x}_t, z_t) \rightarrow  y_t$.

The discriminator receives the current state representation $\ddot{x}_t$ and the generated response map $y_t$ from $G$ and tries to discriminate the actual player responses (real) from generated response maps (fake). We additionally incorporate a classification head $D_{\eta}$ in the discriminator, which learns to output the probabilities for shot types ($\eta_t$). This attaches a supervised objective to the unsupervised objective in the GAN, enabling it to learn from both labelled and unlabelled data. The (Real/ Fake) validation process contains the unsupervised objective of the SS-GAN where it learns the structure and dynamics of the player responses. The classification objective captures the hierarchical relationships among different shot types and determines when the players utilise them, enforcing the generator to learn those player adaptation techniques. Formally the objective of the SS-GAN can be defined as, 

\begin{equation}
\begin{split}
 \mathop{min}_{G} \mathop{max}_{D} \hspace{1mm}\mathbb{E}_{\ddot{x}_t,y_t\sim p_{data}(\ddot{x},y)}[\log D(\ddot{x}_t,y_t)]+ \\ 
 \mathbb{E}_{\ddot{x}_t\sim p_{data}(\ddot{x}),z_t\sim p_{z}(Z)}[\log(1-D(\ddot{x}_t,G(\ddot{x}_t,z_t))]+ \\
  \lambda_{\eta} \mathbb{E}_{\ddot{x}_t,\eta_t\sim p_{data}(\ddot{x}_t,\eta_t)}[\log D_{\eta}(\eta_t|\ddot{x}_t)] ,
\end{split}
\label{eq:semi_supervised}
\end{equation} 

where $D_{\eta}$ is the classification head of D and $\lambda_{\eta}$ is a hyper parameter which controls the contributions of the supervised and unsupervised objectives. 

\subsection{Player Specific Adaptation}
\label{sec:player_specific_adaptation_sec}

Neurological research on human EM has revealed that the EM stores one's own personal experiences rather than the acquired knowledge from other peoples' experiences \cite{tulving1985elements}. This theory is further established by neuroscience research related to tennis shot prediction, where the researches have observed a positive correlation between active practice and performance, and not passive practice \cite{cacioppo2014intention}. 

\begin{figure*}[htbp]
\centering
{\includegraphics[width = .95 \linewidth]{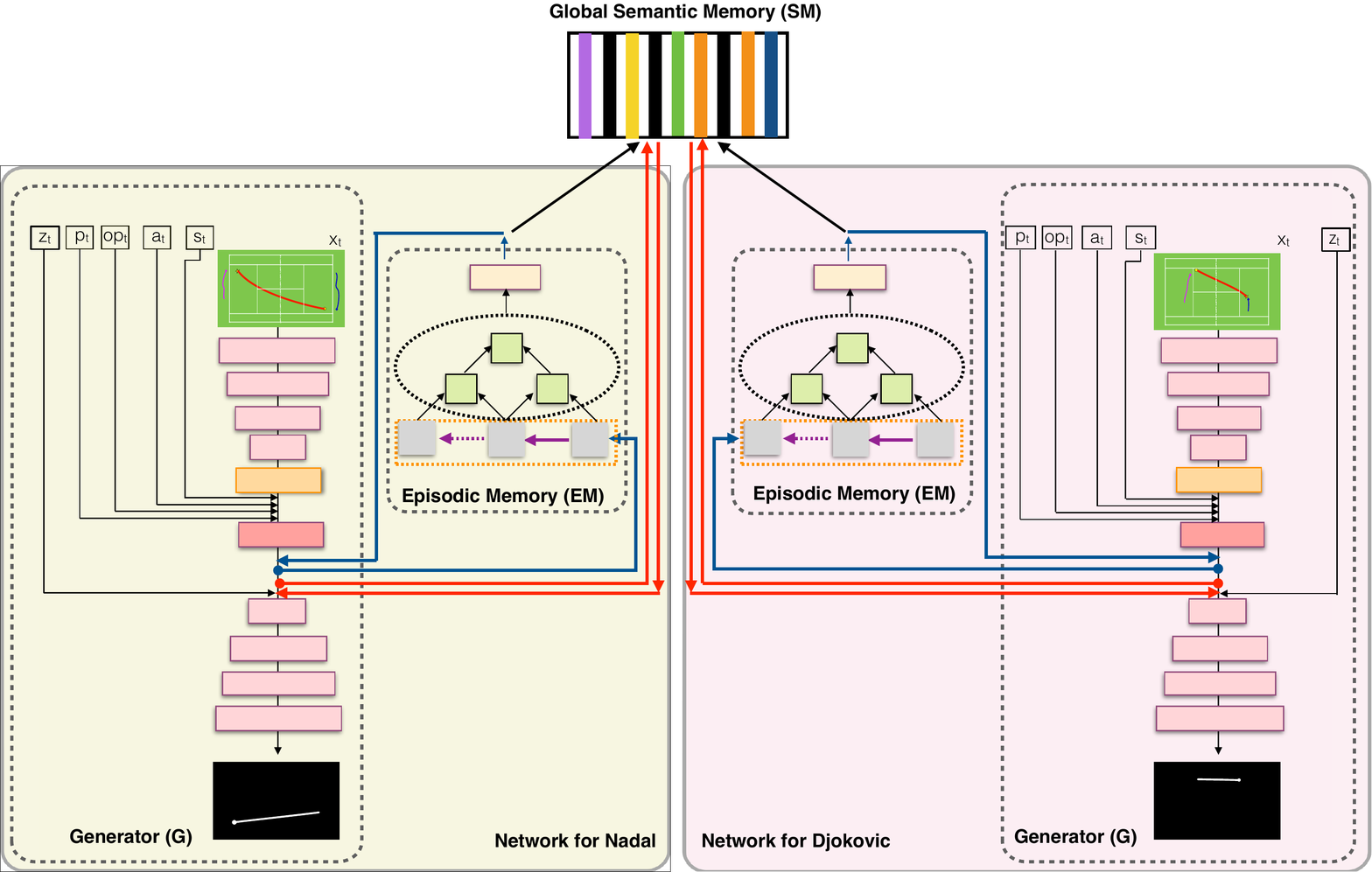}}
\caption{Each player model is composed of a player specific PN, EM and RGN however SM is shared between the players. For the clarity of the illustration we demonstrate the model for only a  2 player scenario, however it could be directly extended to any number of players.}
\label{fig::fig_dual_model}
\end{figure*}

Hence, as the experience of different players depends on the tennis games that they have played, it is not ideal to model EM as a global memory module. Therefore we maintain a separate EM for each player. However it has been shown that SM incorporates extracted knowledge of a cohort of people \cite{wang2017semantic}. When drawing parallels to a tennis scenario SM would represent the rules of the game, the game dynamics, etc. Hence we maintain a global SM for all players. Furthermore, players pay varied levels of attention towards different input features. For instance one player may pay more attention to opponent location where as another may pay more attention to ball incident speed and angle. Hence it is vital to maintain player specific PNs and RGNs. Fig. \ref{fig::fig_dual_model} illustrates the local and global memory architecture that we propose, enabling player specific adaptations. 

\section{Evaluations and Discussion}
\label{sec:evaluations}
\subsection{Dataset}
We used tennis tracking data from the Hawk-eye system for an entire tournament of the 2012 Australian Open Men's singles. The system records (x, y, z) positions of the ball as a function of time, along with the player feet positions at millisecond granularity, and other meta data including current points, time duration, sever and receiver. The dataset consists of around 10,000 shots, however as the tournament progresses in a knock-out format, similar to \cite{wei2016forecasting} we focus our analysis on the top 3 players.

When training the respective player models, we maintain the chronological order of the inputs from the start of the tournament; every shot that has occurred in a game the player has played is fed to the model in this order. Therefore we retain the order of experiences that have occurred, allowing the episodic and semantic memories to replicate the player's brain activities.

\subsection{Shot Type Prediction}
In this experiment we evaluate the performance of the proposed method when predicting the outcome of the next shot, where the model predicts whether the next shot is either a winner, an error or a return shot. The details of the shots played by each player are given in Tab. \ref{tab:shot_details}. For each player, we utilise 70\%, 25\% and 5\% of shots, chronologically,  for training, testing and validation.

\begin{table}[!htb]
  \caption{Counts for different shot types played by the top 3 players in our dataset}
  \label{tab:shot_details}
  \centering
  \begin{tabular}{|ccccc|}
    \hline
    \textbf{Player} & \textbf{Total Shots} &\textbf{Winner} & \textbf{Error} & \textbf{Return} \\
    \hline
   Djokovic & 3,410 & 378 & 554 & 2,478 \\
     Nadal & 3,488 & 215 & 426 & 2,847 \\
     Federar & 1,882 & 187 & 579 & 1,116 \\
     \hline\hline
     \textbf{Total} & \textbf{8,780} & \textbf{780} & \textbf{1,559} & \textbf{6,441} \\
     \hline
\end{tabular}
	\vspace{-2mm}
\end{table}

\subsubsection{Baselines}
As the first baseline model we utilise the Dynamic Bayesian Network (DBN) model proposed in \cite{wei2016forecasting}. This model utilises speed, ball bounce location, player feet locations and a set of hand crafted dominance features to classify the return shot type. In the next baseline we adapt the classifier model proposed in \cite{draschkowitz2014predicting} for classifying the shot success in table tennis matches. Essentially the model utilises hits (number of shots that the player of interest has played thus far from the beginning of the rally), a shot quality variable computed by speed and bounce position of the incoming shot, shot direction, player ids, and current player points. Then we classify these features using a SVM classifier. 

To provide a fair comparison against deep learning models, we pass ball and player trajectories through an LSTM model \cite{hochreiter1997long} and generate the respective classification by passing the LSTM  embeddings through softmax classification function \cite{duan2003multi}. 

\subsubsection{Validation}
Similar to \cite{wei2016forecasting} for all the models we measure the area under the ROC curve (AUC) to assess the performance. Tab. \ref{tab:shot_classification}  presents our evaluation results. 

\begin{table*}[htbp]
  \caption{Shot Type Classification Results: We measure the area under the ROC curve (AUC) to assess the performance. NA stands for Not Available as the metric is not evaluated in that baseline method }
  \label{tab:shot_classification}
  \centering
  \begin{tabular}{|cccc|}
    \hline
    \textbf{Method} & \textbf{AUC- Winner Shots} & \textbf{AUC- Error Shots} & \textbf{AUC- Return Shots} \\
    \hline
   Draschkowitz et. al \cite{draschkowitz2014predicting}  & 52.45  & 61.33& 61.89\\
     LSTM model  &64.61& 66.60 &72.69 \\
     Wei et. al  \cite{wei2016forecasting}&  71.60 & 77.03 & NA\\
     \hline\hline
     \textbf{MSS-GAN} & \textbf{82.65} & \textbf{88.33} &\textbf{89.01}  \\
     \hline
\end{tabular}
	\vspace{-2mm}
\end{table*}

We observe the poorest performance from \cite{draschkowitz2014predicting} as it doesn't posses any capacity to oversee player or scene specific context. The model neither incorporates historical player behaviour nor the ball trajectory information when predicting the shot outcome. The baseline LSTM model incorporates this information, and gains a significant performance boost compared to \cite{draschkowitz2014predicting}. We would like to compare it against the model of Wei et. al \cite{wei2016forecasting}, where the former model has the attained the classification process through an automatic feature learning method. Comparatively similar accuracies emphasises the importance of the hierarchical feature learning process of deep learning models, which automatically learn semantic correspondences through back propagation.  The model proposed by Wei et. al captures the current context through hand crafted player specific and game specific context. In contrast, the proposed model learns these attributes automatically via modelling the player knowledge and experiences through neural memory networks and outperforms the state-of-the-art baselines. 

\subsection{Shot Location Prediction}
\label{sec:shot_location_prediction_sec}
In this experiment we test the performance of the proposed next shot location prediction method against the state-of-the-art baselines. Similar to the previous experiment, for each player, we utilise 70\%, 25\% and 5\% of shots, chronologically, for training, testing and validation.

\subsubsection{Baselines}
As the first baseline model we incorporate the continuous shot location prediction model of \cite{wei2016forecasting}. As the next baseline we utilised the trajectory prediction method of \cite{kumar20113d}. We adapted the system of \cite{kumar20113d} where they try to predict the future ball trajectory from the observed trajectory. Inspired by the encouraging results obtained in \cite{jansson2017predicting} for golf shot prediction, we model the same features of \cite{kumar20113d} using the method of \cite{jansson2017predicting}. The model proposed in \cite{jansson2017predicting} is a deep LSTM model which maps sequential hierarchical relationships among the input features. 

\subsubsection{Validation}
For all the methods we measure the Euclidian distance between the predicted and ground truth locations in meters as the prediction error. Tab. \ref{tab:shot_location_prediction} presents the performance of the proposed method against the 3 state-of-the-art baselines. As \cite{kumar20113d} utilises a simple physical model to predict the future ball trajectory, assuming that the ball follows a parabolic arc under gravitational force, we observe poor performance from it. The model in \cite{jansson2017predicting} improves upon this via hierarchical feature learning from the ball trajectory characteristics through a recurrent model.

The method of \cite{wei2016forecasting} builds upon the trajectory features, via incorporating the game and player context into the prediction framework, however, fails to capture salient information from longterm dependencies among player behavioural patterns. In contrast, we capture those hierarchically, allowing us to effectively propagate this information into the future action generation pipeline. 


\begin{table*}[htbp]
 \caption{Shot Location Prediction Results: We measured the distance between the predicted and ground truth shot locations in meters and report the Average ($\mu$) and the Standard Deviation of this distances ($\sigma$)}
  \label{tab:shot_location_prediction}
 \centering
\begin{tabular}{|c|c|c|c|c|c|c|c|c|}
\hline
\multirow{2}{*}{Method} & \multicolumn{2}{c|}{Nadal} & \multicolumn{2}{c|}{Djokovic} & \multicolumn{2}{c|}{Federer} & \multicolumn{2}{c|}{Overall} \\ \cline{2-9} 
                        & $\mu$          & $\sigma$         & $\mu$            & $\sigma$          & $\mu$           & $\sigma$          & $\mu$           & $\sigma$          \\ \hline
Kumar et. al   \cite{kumar20113d}          & 4.23         &   1.0043         & 3.87           &     0.9145           & 5.83          &   2.1534           & 4.64          &   1.357           \\ \hline
Jansson et al.  \cite{jansson2017predicting}         & 2.11         &    0.8114         & 1.95           &   0.7671           & 3.41          &    0.9833          & 2.49          &  0.8539          \\ \hline
Wei et. al        \cite{wei2016forecasting}       & 1.72         &   0.5430      & 1.64           &    0.3034          & 2.32          &      0.7016        & 1.89          &      0.5160        \\ \hline
\textbf{MSS-GAN}               & \textbf{0.87}         &     \textbf{0.0302}       & \textbf{0.79}           &  \textbf{0.0210 }          & \textbf{1.14}          &   \textbf{0.0330}            & \textbf{0.93}          & \textbf{0.0280}            \\ \hline
\end{tabular}
\end{table*}

\subsection{Ablation Experiments}
To further demonstrate our proposed approach, we conducted a series of ablation experiments, identifying the crucial components of the proposed methodology. In the same setting as Sec. \ref{sec:shot_location_prediction_sec}, we evaluate the proposed MSS-GAN model against a series of counterparts constructed by removing strategic components of the MSS-GAN as follows:
\begin{enumerate}
\item $G^{G}/(D, M^{EM}, M^{SM})$: removes the discriminator, $M^{EM}$ and $M^{SM}$ models and is trained with the supervised learning objective.
\item $(G, D)^{G}/(M^{EM}, M^{SM})$: removes the $M^{EM}$ and $M^{SM}$ models and is trained with the GAN objective of \cite{isola2017image}.
\item $(G, D)^{G,*}/(M^{EM}, M^{SM})$ Similar to the method 2, however it is trained with the semi supervised objective defined in Eq. \ref{eq:semi_supervised}. 
\item $(G, D, M^{EM})^{G}/( M^{SM})$: removes the $M^{SM}$ and is trained globally without the player adaptation techniques of Sec. \ref{sec:player_specific_adaptation_sec}.
\item $(G, D, M^{EM})^{L}/( M^{SM})$:  Similar to the previous model however uses the player adaptation techniques of Sec. \ref{sec:player_specific_adaptation_sec}.
\item $(MSS-GAN)^{G}$: Same as the proposed MSS-GAN model, however we train one global model without the player adaptation technique. 
\end{enumerate}

\begin{table*}[htbp]
  \caption{Ablation Experiment Results: We measured the distance between the predicted and ground truth shot locations in meters.}
  \label{tab:shot_location_prediction}
  \centering
  \begin{tabular}{|ccccc|}
    \hline
    \textbf{Method} & \textbf{Nadal} & \textbf{Djokovic} & \textbf{Federer} & \textbf{Average} \\
    \hline
     $G^{G}/(D, M^{EM}, M^{SM})$ & 2.03 & 1.98 &3.88 &  2.63 \\
     $(G, D)^{G}/(M^{EM}, M^{SM})$ & 1.63 & 1.59 &2.18 & 1.80 \\
     $(G, D)^{G,*}/(M^{EM}, M^{SM})$& 1.44 & 1.32 &1.95 & 1.57\\
     $(G, D, M^{EM})^{G}/( M^{SM})$ & 1.21 & 1.15 &1.44 & 1.26 \\
     $(G, D, M^{EM})^{L}/( M^{SM})$& 1.13 & 1.10 &1.36 &1.19 \\
     $(MSS-GAN)^{G}$& 1.02 & 1.03 &1.23 & 1.09\\
     \hline\hline
     \textbf{MSS-GAN} &\textbf{0.87} & \textbf{0.79} & \textbf{1.14}  & \textbf{0.93} \\
     \hline
\end{tabular}
	\vspace{-2mm}
\end{table*}

We observe the lowest accuracy from the $G^{G}/(D, M^{EM}, M^{SM})$ model which fails to capture the context and the dynamics of the tennis game through an off the shelf supervised learning loss function. We observe a significant performance boost with the unsupervised GAN objective. However it lacks the capacity to capture salient information from the historical embeddings, and as such the performance increases from $(G, D)^{G}/(M^{EM}, M^{SM})$ to  $(G, D, M^{EM})^{G}/( M^{SM})$ where the latter has further capacity to understand the player level behavioural differences. We would like to compare the performance of the $MSS-GAN^G$ model against the proposed $MSS-GAN$ model, where the former model learns one single network for all the players. The different in accuracies are significant, emphasising the importance of capturing player level semantics separately.  We further compare models $(G, D)^{G}/(M^{EM}, M^{SM})$ against  $(G, D)^{G,*}/(M^{EM}, M^{SM})$. Similar to observations of \cite{odena2016semi,denton2016semi}, our results demonstrate the importance of a semi supervised learning objective, which compliments the generator in learning the hierarchical attributes of the scene.   

\subsection{Qualitative Evaluations}

\begin{figure}[htbp]
\centering
\subfigure[]{\includegraphics[width = .3 \linewidth]{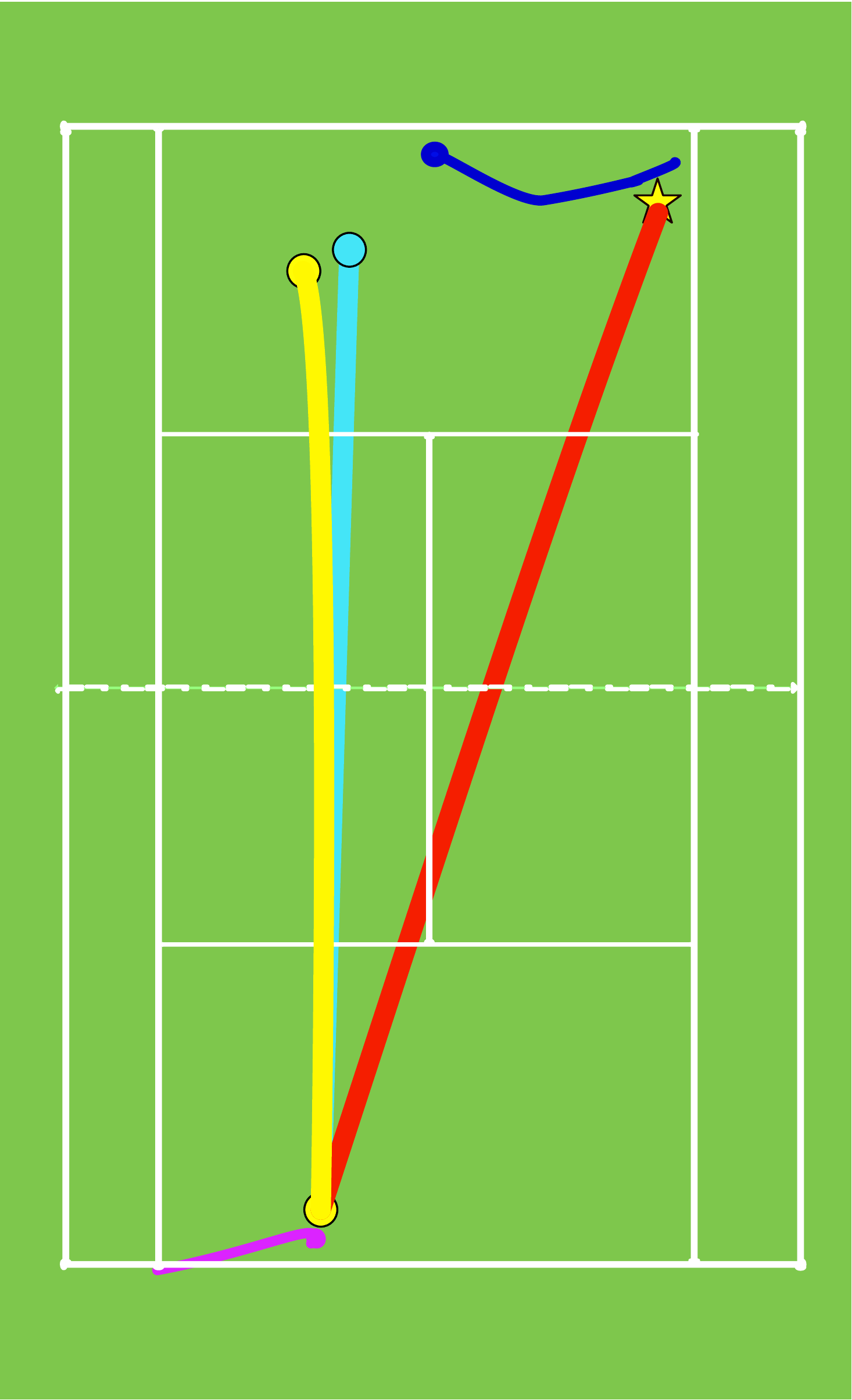}} 
\subfigure[]{\includegraphics[width = .3 \linewidth]{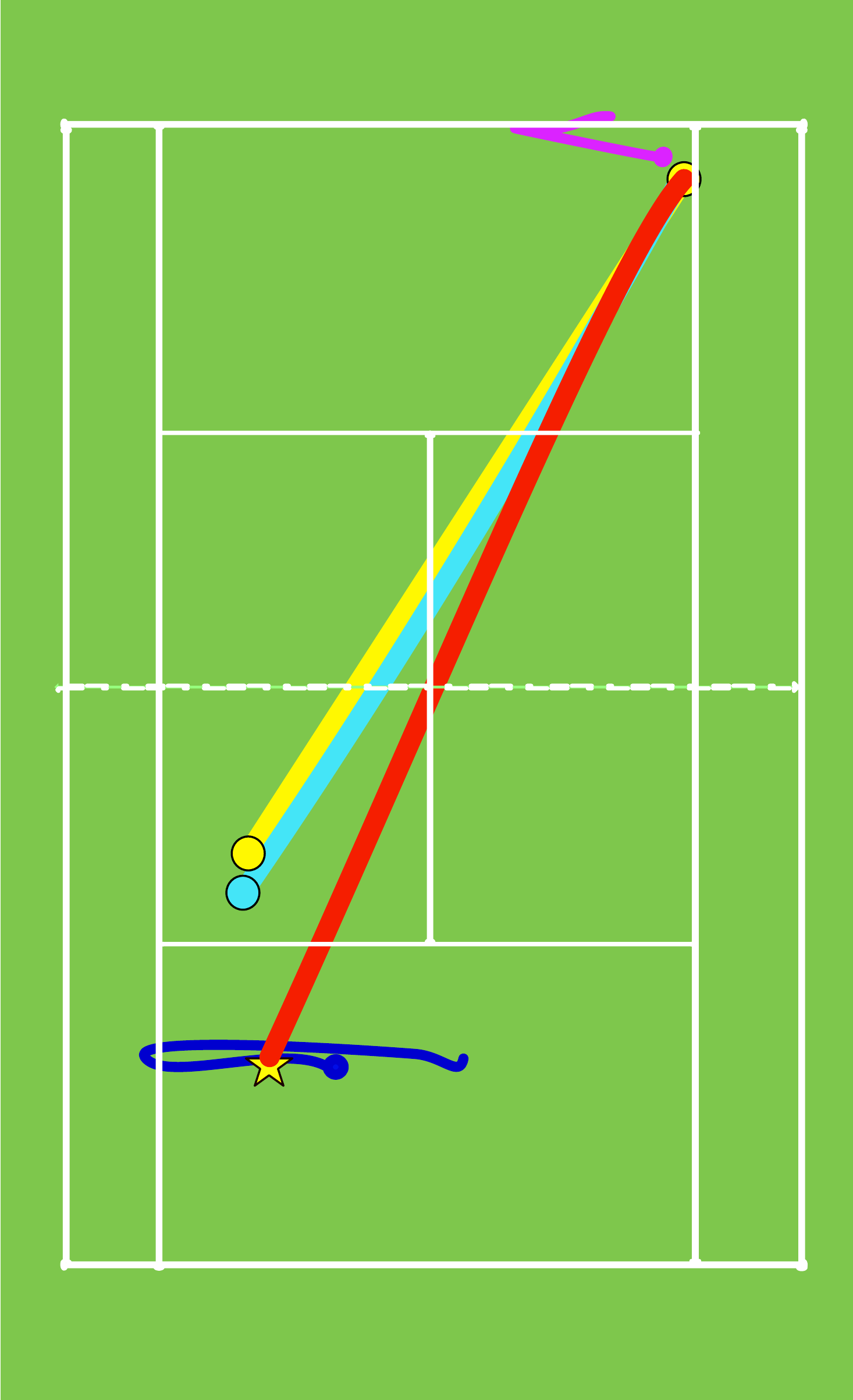}} 
\subfigure[]{\includegraphics[width = .3 \linewidth]{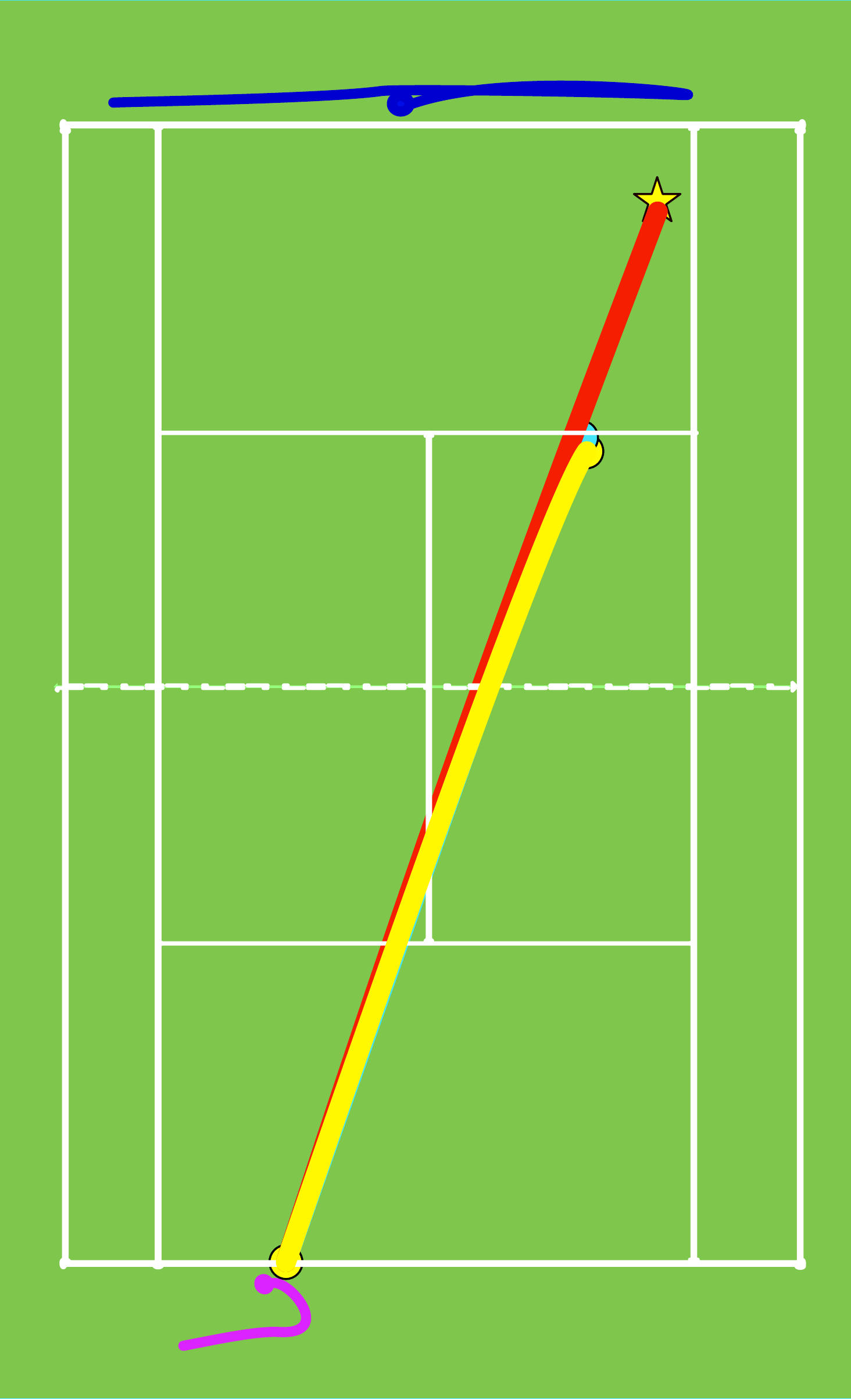}} 
\subfigure[]{\includegraphics[width = .3 \linewidth]{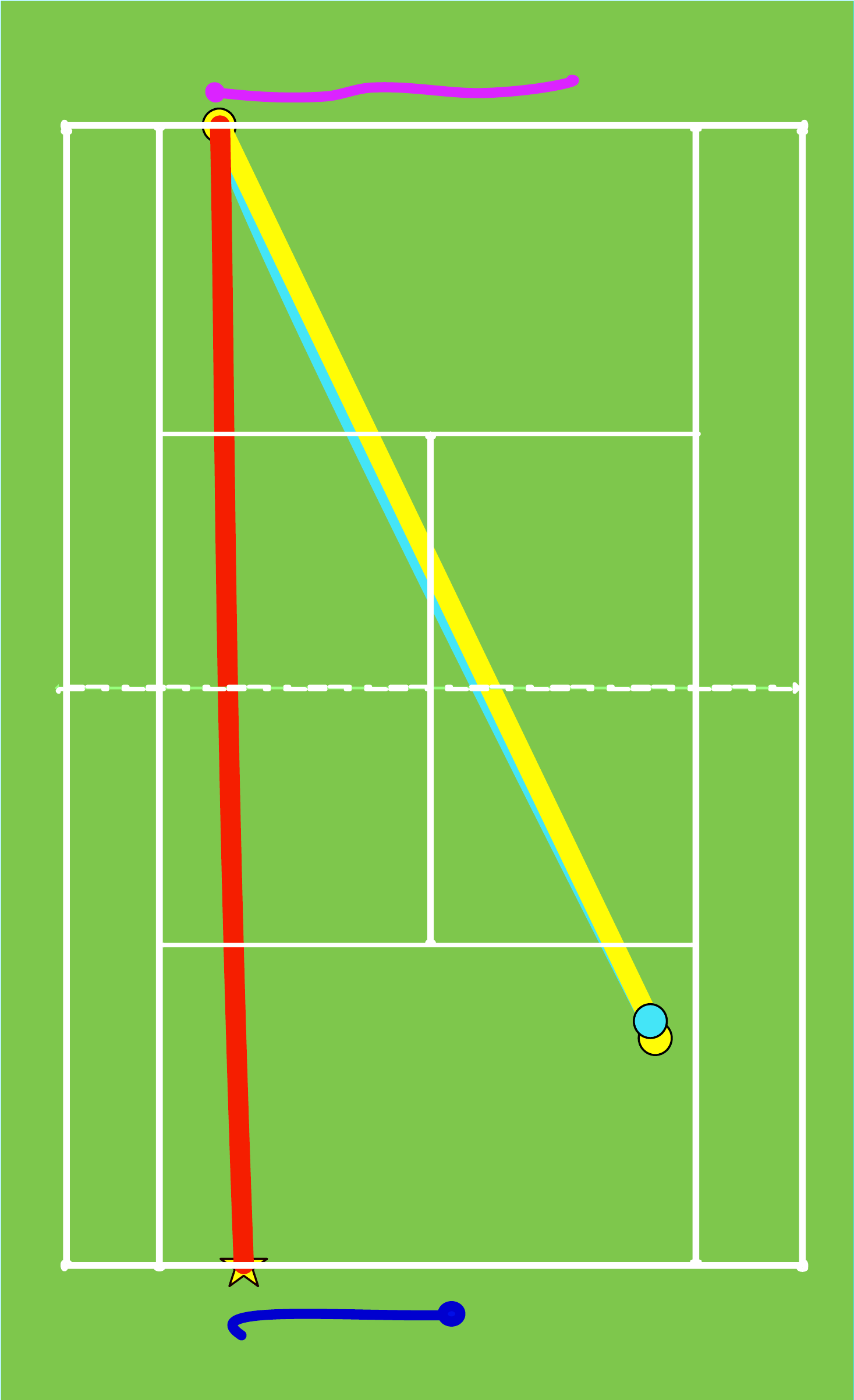}} 
\subfigure[]{\includegraphics[width = .3 \linewidth]{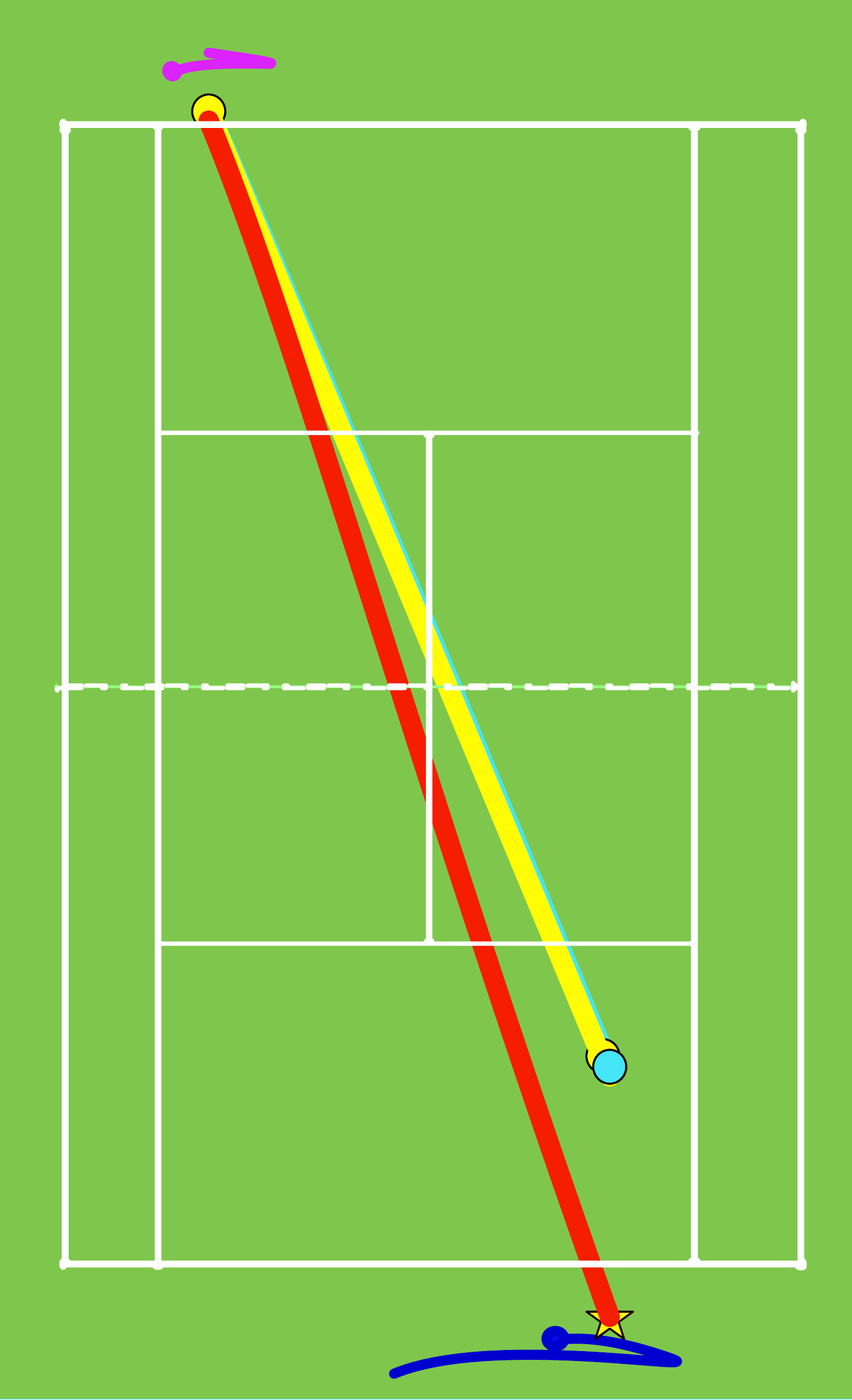}} 
\subfigure[]{\includegraphics[width = .3 \linewidth]{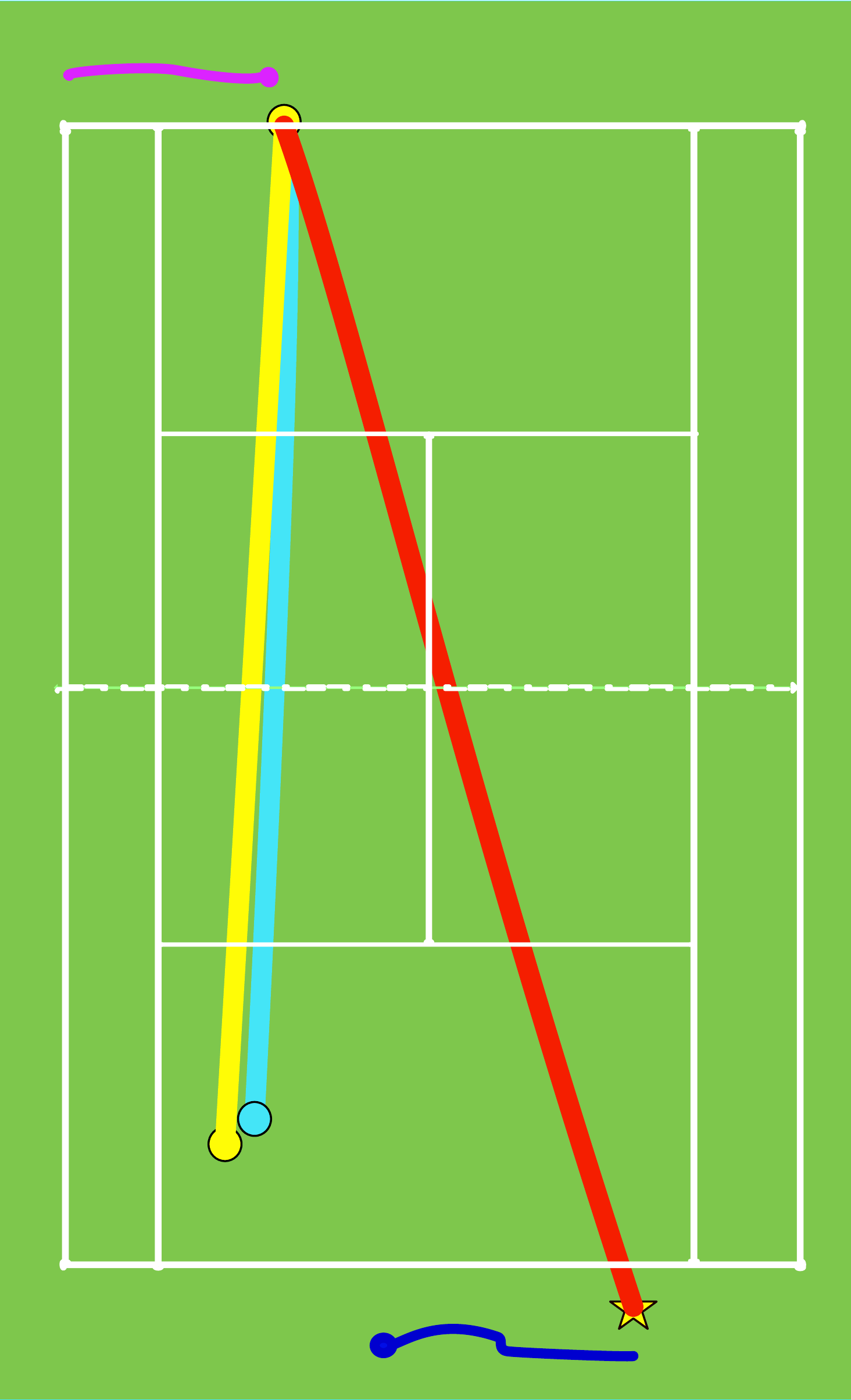}} 
\subfigure[]{\includegraphics[width = .3 \linewidth]{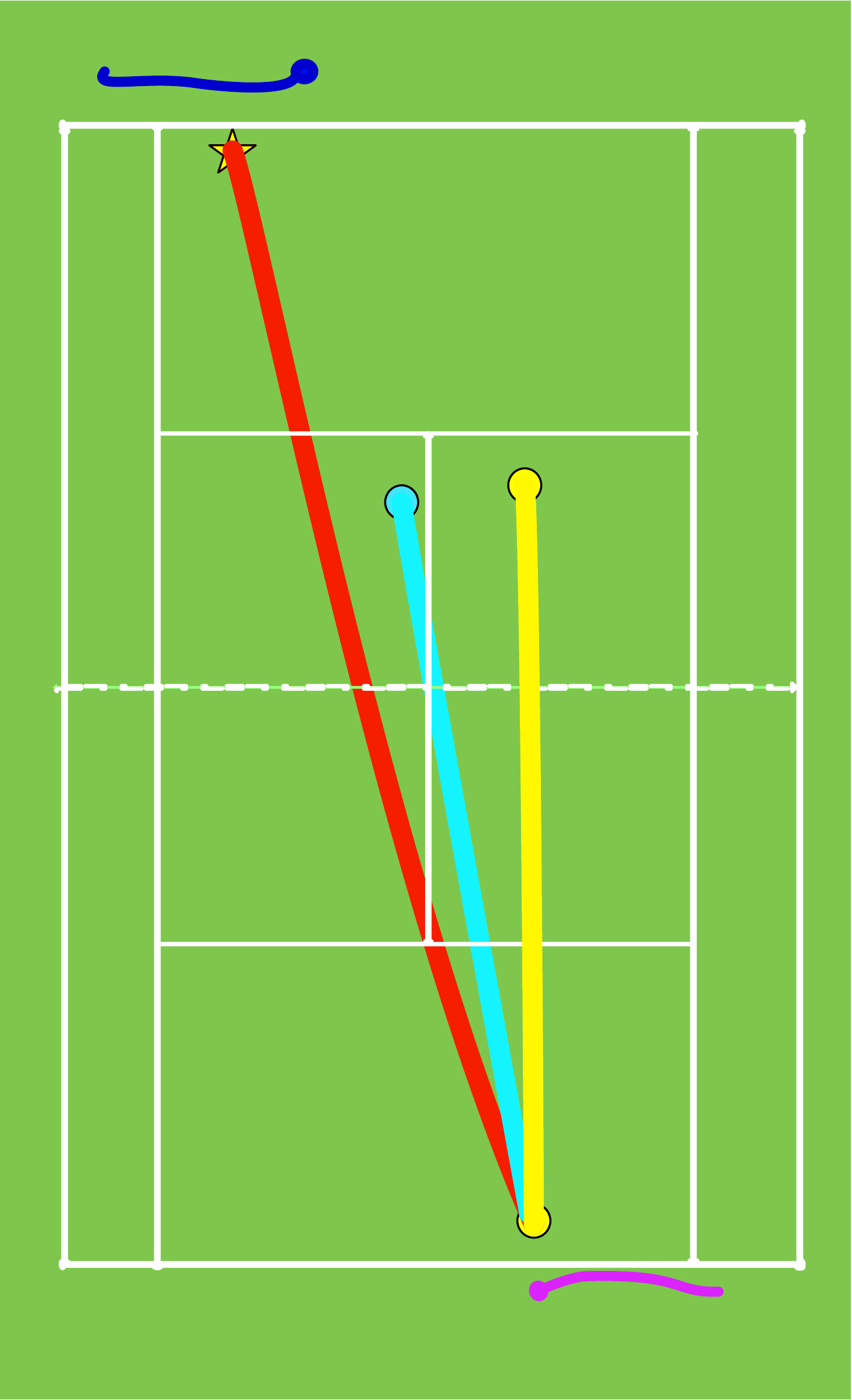}} 
\subfigure[]{\includegraphics[width = .3 \linewidth]{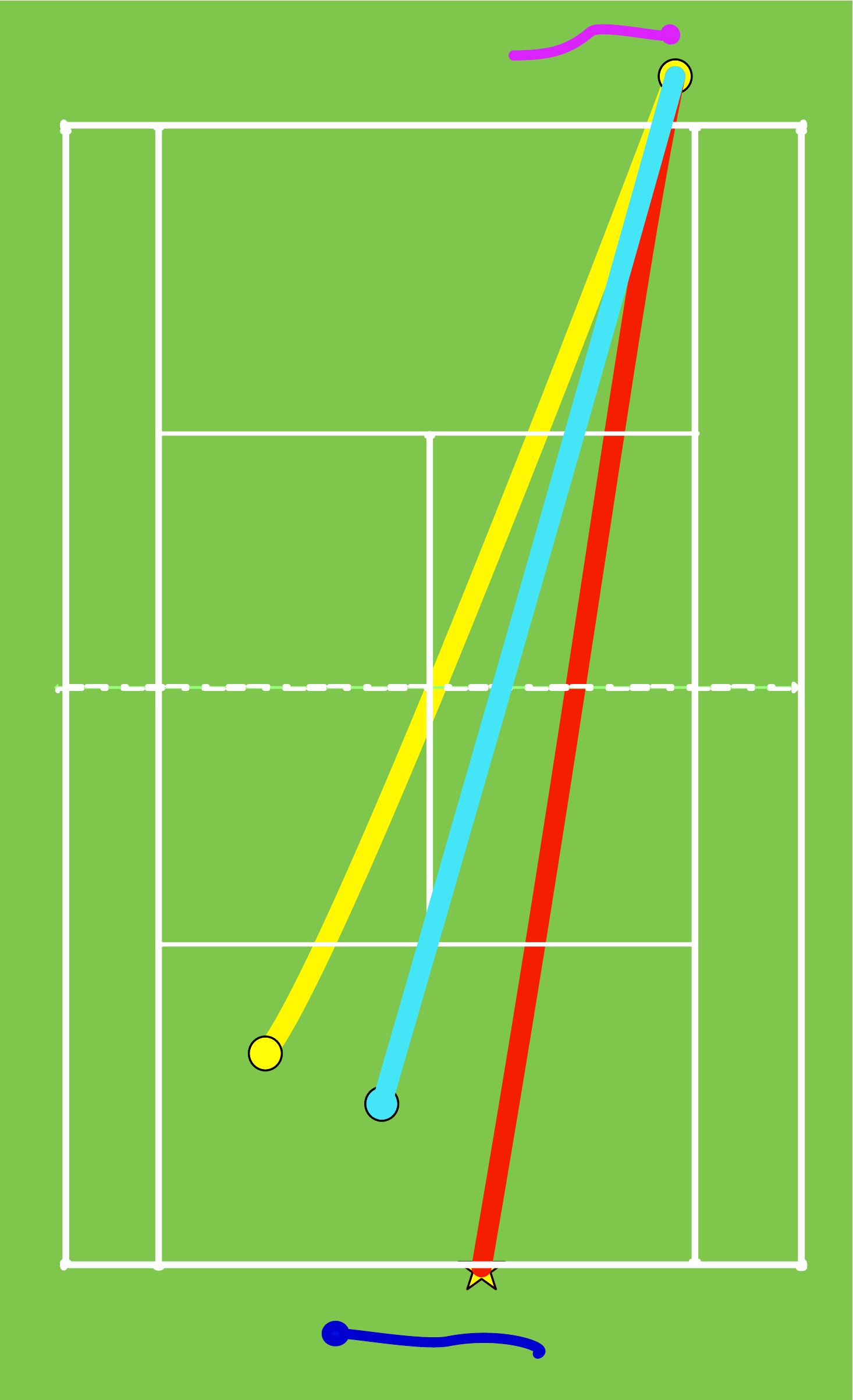}} 
\subfigure[]{\includegraphics[width = .3 \linewidth]{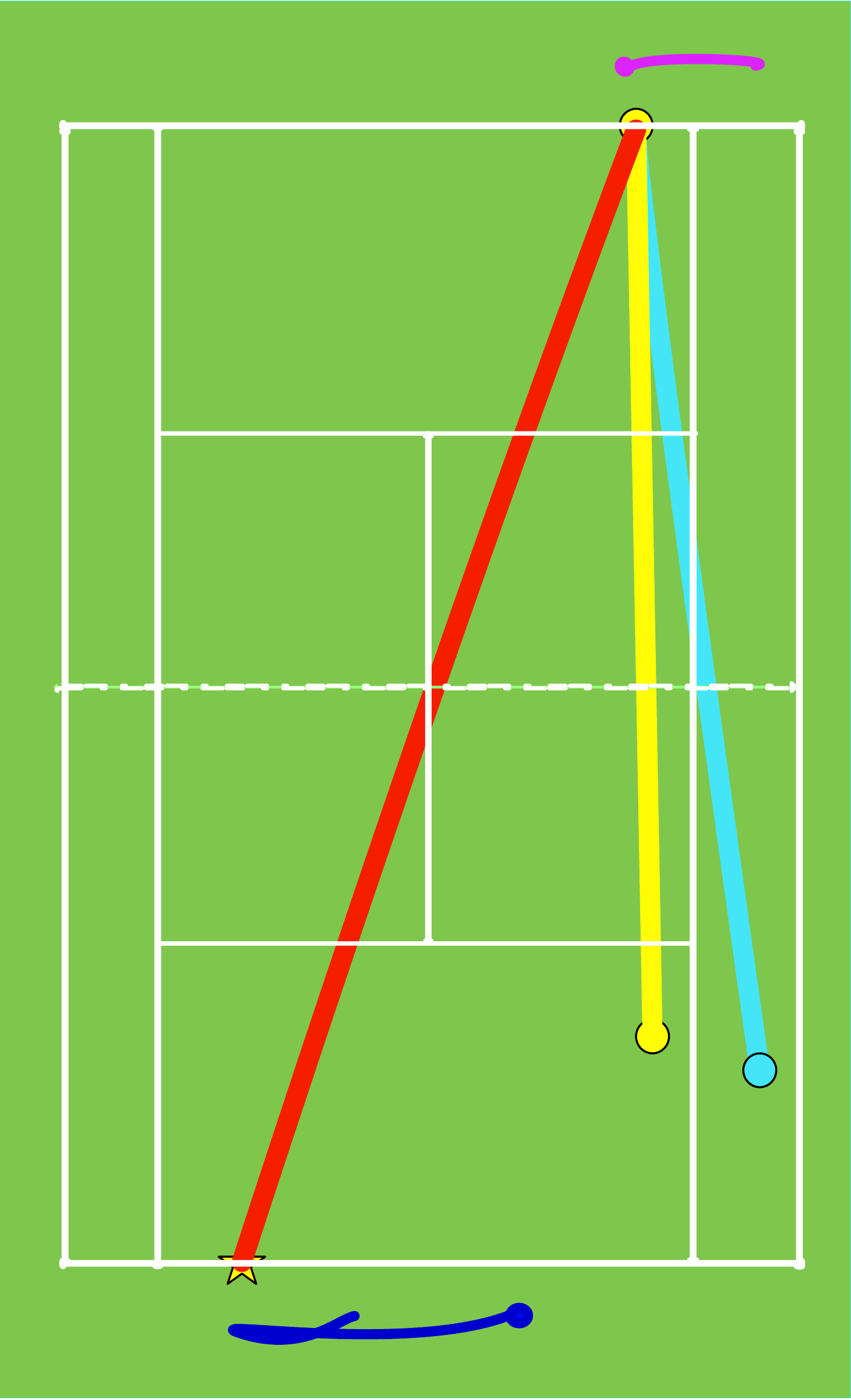}} 
\caption{Qualitative results from the proposed MSS-GAN model. Incoming shot is in red where the yellow star and the circle denotes the shot starting and ending locations. Feet movements of player of interest and the opponent are in magenta and blue colours. Ground truth and predicted trajectories are denoted in cyan and yellow lines, respectively.First two rows show accurate predictions while the 3rd row shows scenarios where the predicted trajectory deviates from the ground truths. However in (g) and (h) we observe that the predicted trajectory maximises the opportunity of the winning probability of the player of interest. Please note that we have overlaid the predictions from RGN on top of court outlines for the clarity of illustration}
\label{fig:qualitative}
\end{figure}

Qualitative evaluations from the proposed MSS-GAN model are presented in Fig. \ref{fig:qualitative}. We denote the incoming shot in red where the yellow star and the circle denotes the shot starting and ending locations. The feet movements of the player of interest and the opponent are denoted in magenta and blue. Ground truth and predicted trajectories are denoted with cyan and yellow lines, respectively.

In the first two rows we have presented accurate predictions while in the last row we observe some deviations from the ground truth. However they are all possible next shot locations for the incoming shots, and for scenarios such as Fig. \ref{fig:qualitative} (h) and (i) we observe that the predicted shot maximises the wining probability for the player of interest compared to the actual ground truth trajectory. 

\begin{figure*}[htbp]
\centering
{\includegraphics[width = .95 \linewidth]{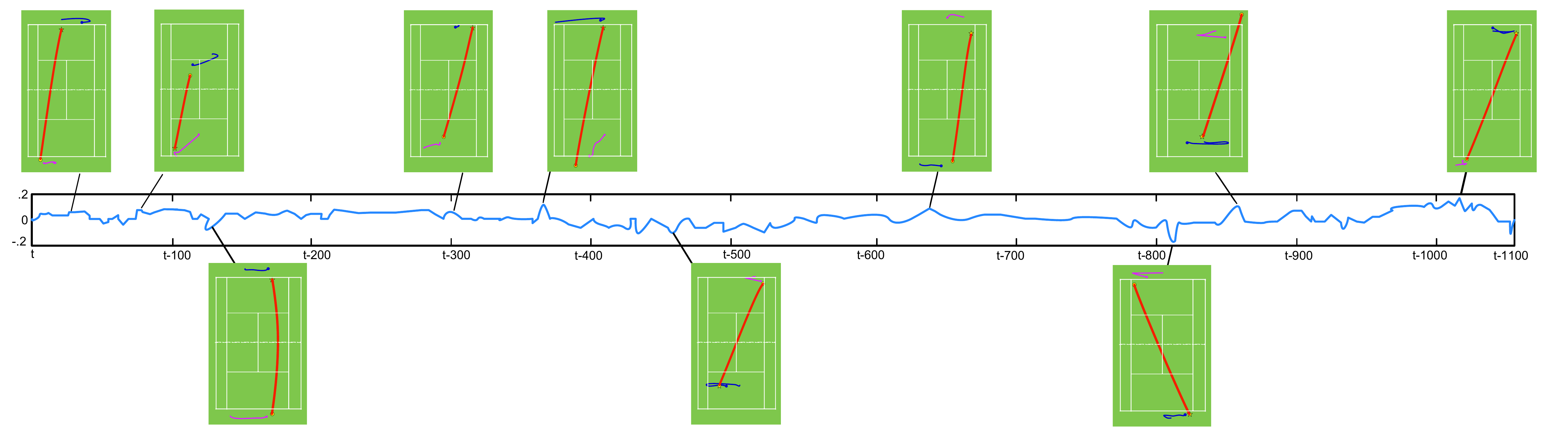}}
\caption{Distribution of memory activations from the first layer of the EM module for the observed shot in Fig. \ref{fig:qualitative} (a). This layer contains 1,000 memory slots which are denoted t to t -1100 indicating the history that has been observed. For different peaks and valleys of the memory activations we also show what the memory has seen at those particular time steps. The model generates higher activations for similar shot patterns, and activations closer to zero for cases where the player has observed different experiences. We effectively propagate this information from the first layer of the memory to the top most layer via combining the salient information in a hierarchical manner.}
\label{fig:fig_EM_activations}
\end{figure*}

Fig. \ref{fig:fig_EM_activations} shows the distribution of activations from the first layer of the proposed EM tree for the Djokovic model, for the observed shot trajectory given in Fig. \ref{fig:qualitative} (a). As N = 1,100, there exist 1,100 memory slots in this layer, hence the historical embeddings range from t to $t - 1100$. For different peaks and valleys in the memory activations, we show what the model has seen at that particular time step.

We observe higher responses for recent events as well as for similar shot patterns in the long term history. It can be clearly seen that the activation pattern takes the current context into consideration and attends all the previous experiences of the player stored in the memory, in order to determine the optimal way to behave. Hence we observe higher activations to similar shot patterns that reside within the entire history captured by the EM module (see the activation peeks between $t - 300$ to $t - 400$ and $t -800$ to $t- 1100$).

\subsection{Impact of the Training Data}

In this section we investigate the impact of training set size, input image dimensions, and the selection of player identities for extraction of training data, for the performance of the proposed MSS-GAN model. 

\subsubsection{Impact of Training Set Size}

In order to analyse the robustness of the proposed model to different training set sizes we analyse the distribution of the average shot location prediction error and training time for an epoch on a single core of an Intel Xeon E5-2680 2.50GHz CPU, for different training set sizes for the training data defined in Sec. \ref{sec:shot_location_prediction_sec}. For this evaluation we used the same testing and validation sets used in Sec. \ref{sec:shot_location_prediction_sec} which are not used for training. When creating the reduced training set, we take data from the first shot Nadal played in the tournament up to the specified number of samples, i.e. when training with 100 samples we take the first 100 shots, and when training with 1000 samples we take the first 1000 shots.                        

\begin{figure}[htpb]
    \centering
      \includegraphics[width=.9\linewidth]{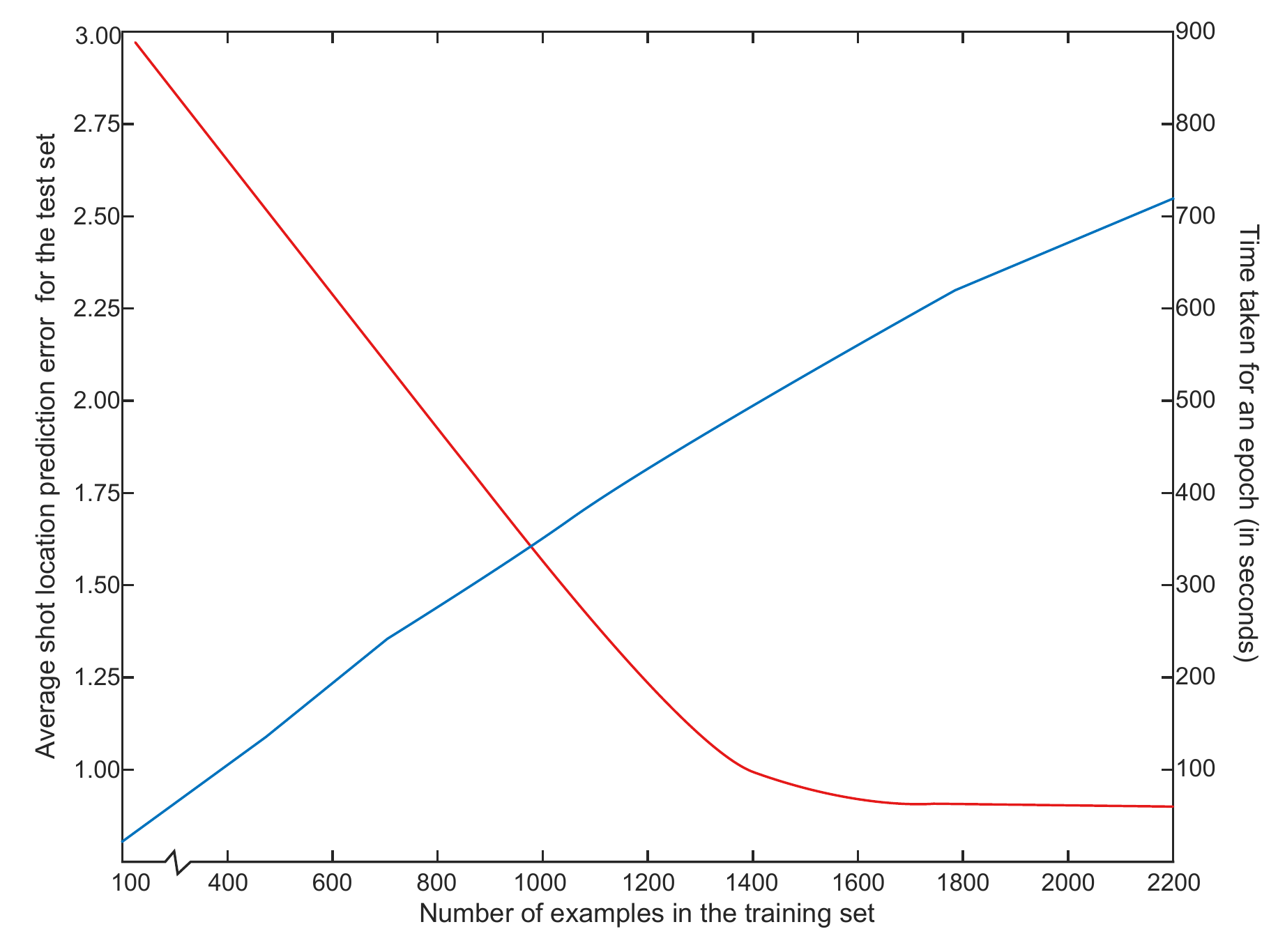} 
     \caption{Change in average shot location prediction error (in meters) and training time for an epoch (in seconds), against training set size}
     \label{fig:traing_set_size_vs_run_time}
 \end{figure}

In Fig. \ref{fig:traing_set_size_vs_run_time} we visualise the average shot location prediction error against different training set sizes (in red) and the elapsed time per epoch (in blue). As could be expected, the training time increases gradually as more samples are added to the corpus. However the model accuracy converges around 1600 training examples and we do not observe substantial improvement, irrespective of the introduction of additional examples. 1600 shots are roughly equivalent to the total number of shots that he has played in first 3 matches.

\subsubsection{Impact of Input Image Size}

In our implementation the input/output image size is set to be 512 x 512 pixels. In order to demonstrate the robustness of the proposed MSS-GAN model to different input image sizes we evaluated different input/outputs sizes of 2048 x 2048, 1024 x 1024, 256 x 256, 128 x128, and 64 x 64 pixels and the evaluated shot location prediction error for the same conditions as in Sec. \ref{sec:shot_location_prediction_sec}. The receptive field sizes of the layers and the number of kernels are kept constant in this experiment. Tab. \ref{tab:shot_location_prediction_input_output_size} shows the average error for the predictions, and we observe that for sizes 1024 x 1024 and 256 x 256 there isn't significant deviation in performance compared to the results of Tab. \ref{tab:shot_location_prediction}. However the performance starts degrading when the input/output sizes are smaller than 256 x 256 pixels. This is due to the granularity of the spatial representation. When the input size is smaller the ground truth trajectory is substantially downscaled to represent it in the image representation. Hence the input and ground truth trajectory representations are less informative to the model. Hence reduction in the granularity of the image representation leads to poor performance. Conversely when the input size it too large it leads to a substantial increase in the number of parameters that require training. Hence the model could not be effectively trained using the limited data available, again leading to poor performance. It should be noted that the same input and output sizes are used for the convenience in the implementation. However this is not a constraint of the proposed MSS-GAN method.

\begin{table}[htbp]
  \caption{Shot Location Prediction Results for Different Input/ Output Image Sizes: We measured the distance between the predicted and ground truth shot locations in meters.}
  \label{tab:shot_location_prediction_input_output_size}
  \centering
  \begin{tabular}{|c|c|}
    \hline
    \textbf{Input/ Output size} &  \textbf{Average} \\
      \hline
     {2048 x 2048} & {1.05}  \\
      {1024 x 1024} & {0.94}  \\
       {512 x 512} & \textbf{0.93}  \\
        {256 x 256 } & \textbf{0.93}  \\
         {128 x 128 } & {1.13}  \\
         {64 x 64 } & {1.24}  \\
  
     \hline
\end{tabular}
	\vspace{-2mm}
\end{table}

\subsubsection{Impact of Selected Players}

Similar to  \cite{wei2016forecasting} we present the main evaluation in Sec. Sec. \ref{sec:evaluations} only for the top-3 players as it allows us to present direct comparisons with \cite{wei2016forecasting}. However in order to better evaluate the predictive power of the proposed MSS-GAN model we randomly selected 4 players from the 2012 Australian Open Men's singles tournament who had progressed up to the fourth round. We trained the model with the data from first 3 rounds and tested on the 4th round. For these players we evaluate the shot location prediction accuracy as it is a more challenging task than predicting the shot type.

\begin{table*}[htbp]
  \caption{Shot Location Prediction Results for Randomly Selected 4 Players from fourth round: We measured the distance between the predicted and ground truth shot locations in meters.}
  \label{tab:shot_location_prediction_2}
  \centering
  \begin{tabular}{|cccccc|}
    \hline
    \textbf{Method} & \textbf{J-W Tsonga} & \textbf{Richard Gasquet} & \textbf{K Nishikori} & \textbf{Andy Murray}& \textbf{Average} \\
     \hline
     Wei et. al  \cite{wei2016forecasting}& 2.13 & 2.60 &3.31 & 2.90 & 2.74\\
     \hline\hline
     \textbf{MSS-GAN} &\textbf{0.97} & \textbf{0.91} & \textbf{1.07} & 1.03 & \textbf{1.00}  \\
     \hline
\end{tabular}
	\vspace{-2mm}
\end{table*}

In order to better appreciate the predictive performance of the proposed MSS-GAN model we also trained Wei et. al's \cite{wei2016forecasting} model on the selected 4 players. When comparing the prediction results presented in Tab. \ref{tab:shot_location_prediction} with Tab. \ref{tab:shot_location_prediction_2} it is clear that the reduced training data has a significant impact on the performance of Wei et. al's \cite{wei2016forecasting} approach. In contrast we do not observe significant deviation in the proposed model's performance, indicating its ability to infer different player styles even without very large volumes of data.

\subsection{Implementation Details}
\begin{figure*}[htbp]
\centering
\subfigure[]{\includegraphics[width = .3 \linewidth]{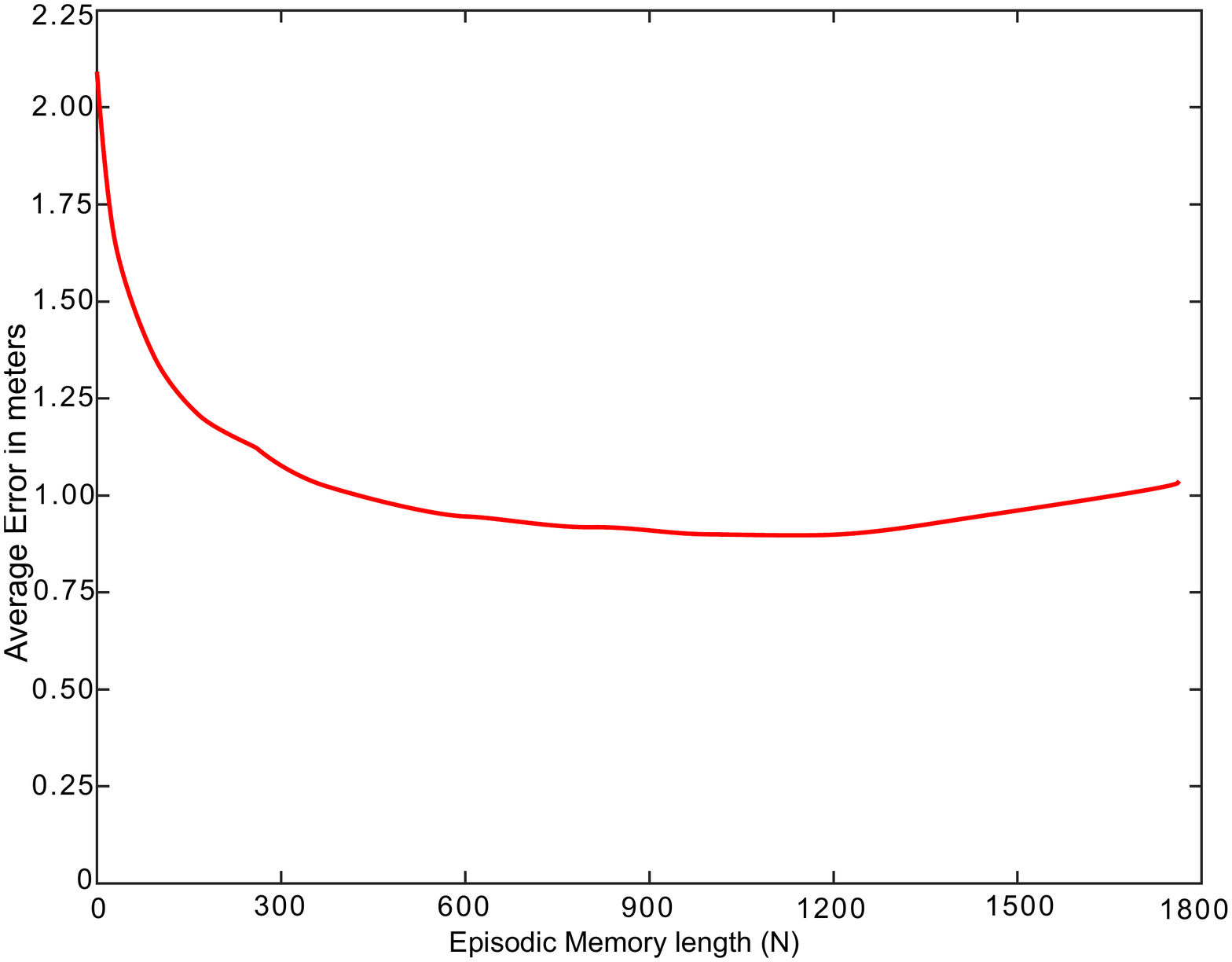}} 
\subfigure[]{\includegraphics[width = .3 \linewidth]{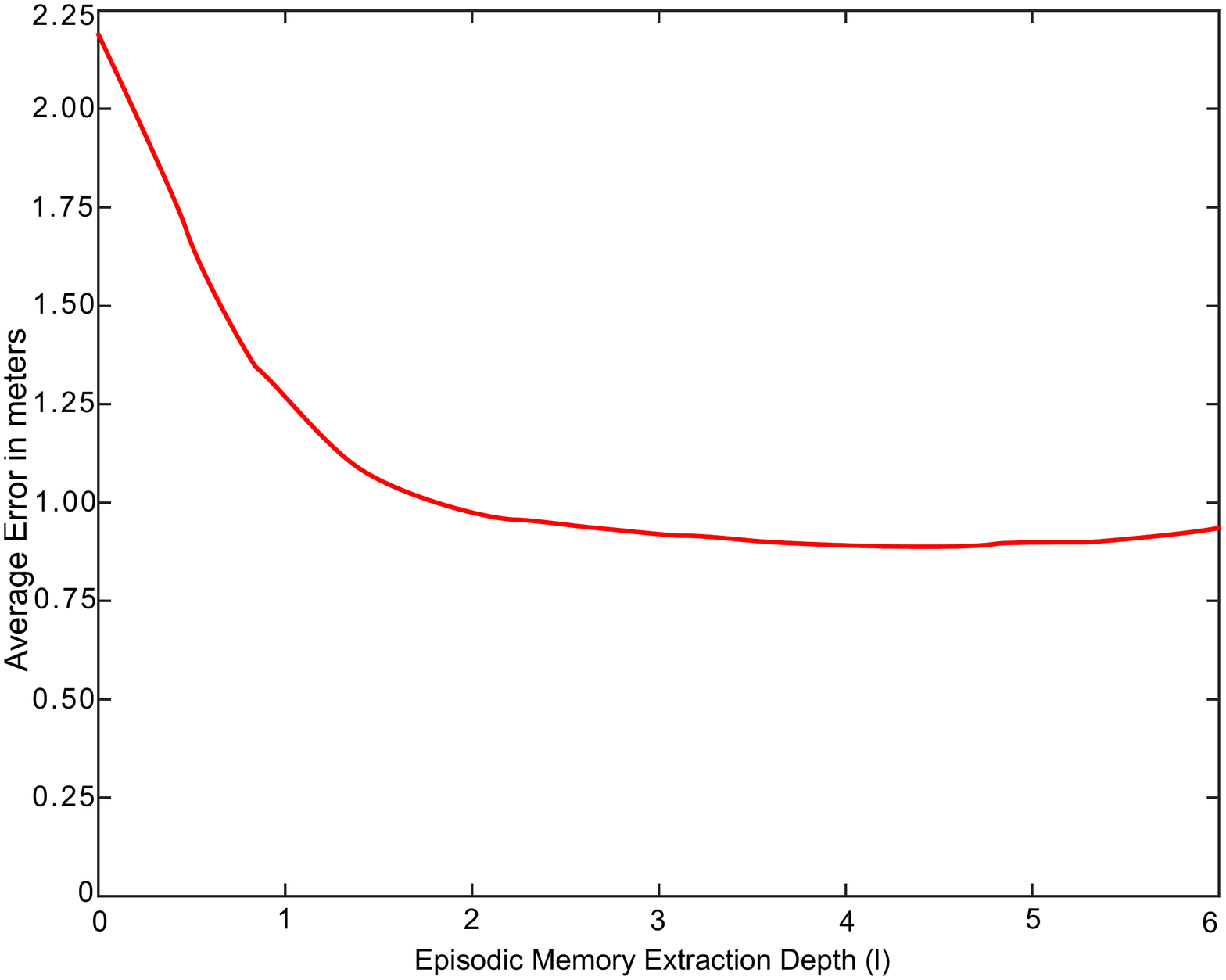}} 
\subfigure[]{\includegraphics[width = .3 \linewidth]{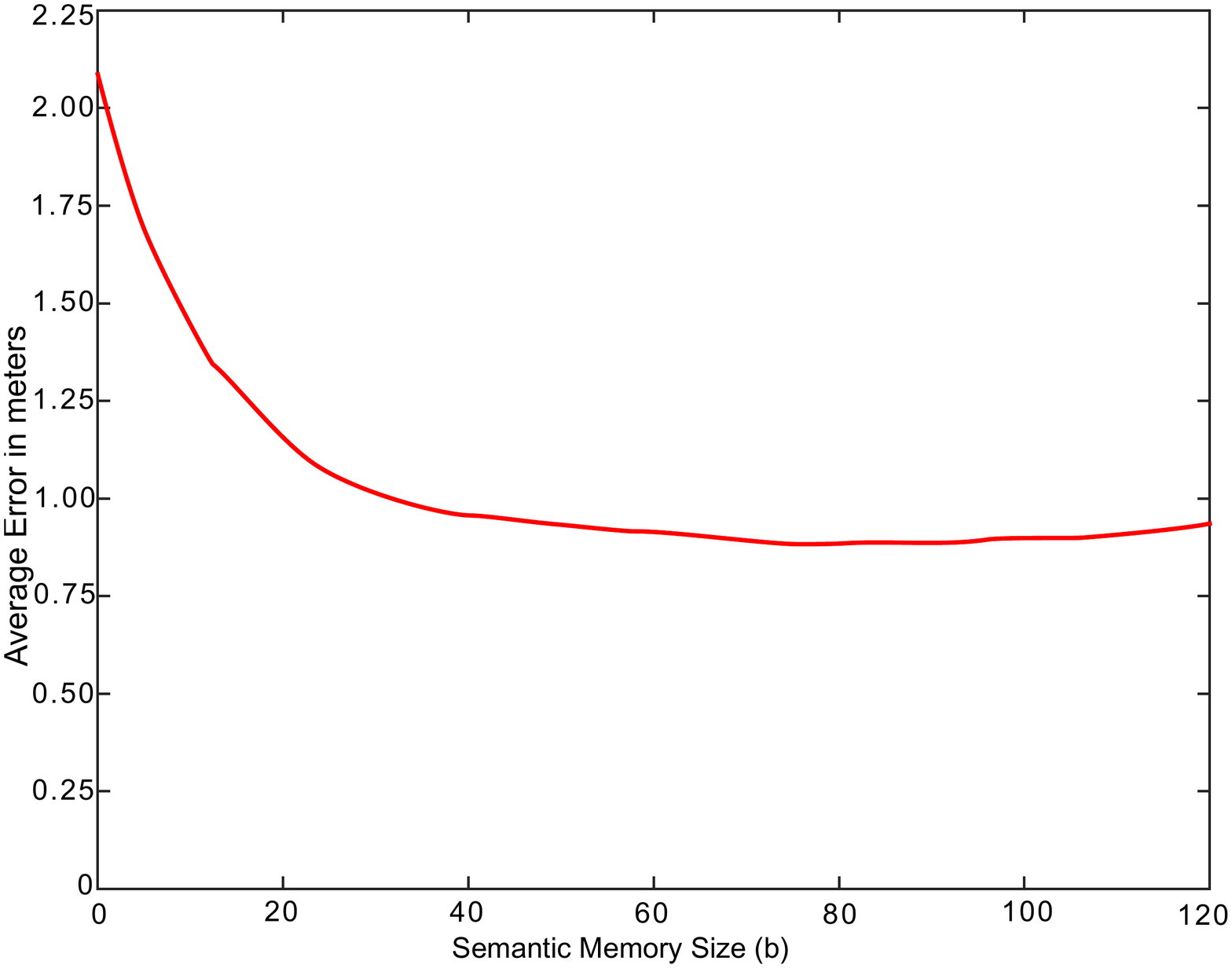}} 
\caption{Hyper parameters evaluation process. We evaluate Episodic Memory length, $N$, Episodic Memory extraction depth, $l$ and Semantic Memory size, $b$, experimentally holding the rest of hyper parameters constant. As $N=1100$, $l=3$ and $b=80$ gives us the optimal results we set the respective sizes accordingly.  }
\label{fig:hyper_parameters}
\end{figure*}

\begin{figure}[htbp]
\centering
\includegraphics[width = .7 \linewidth]{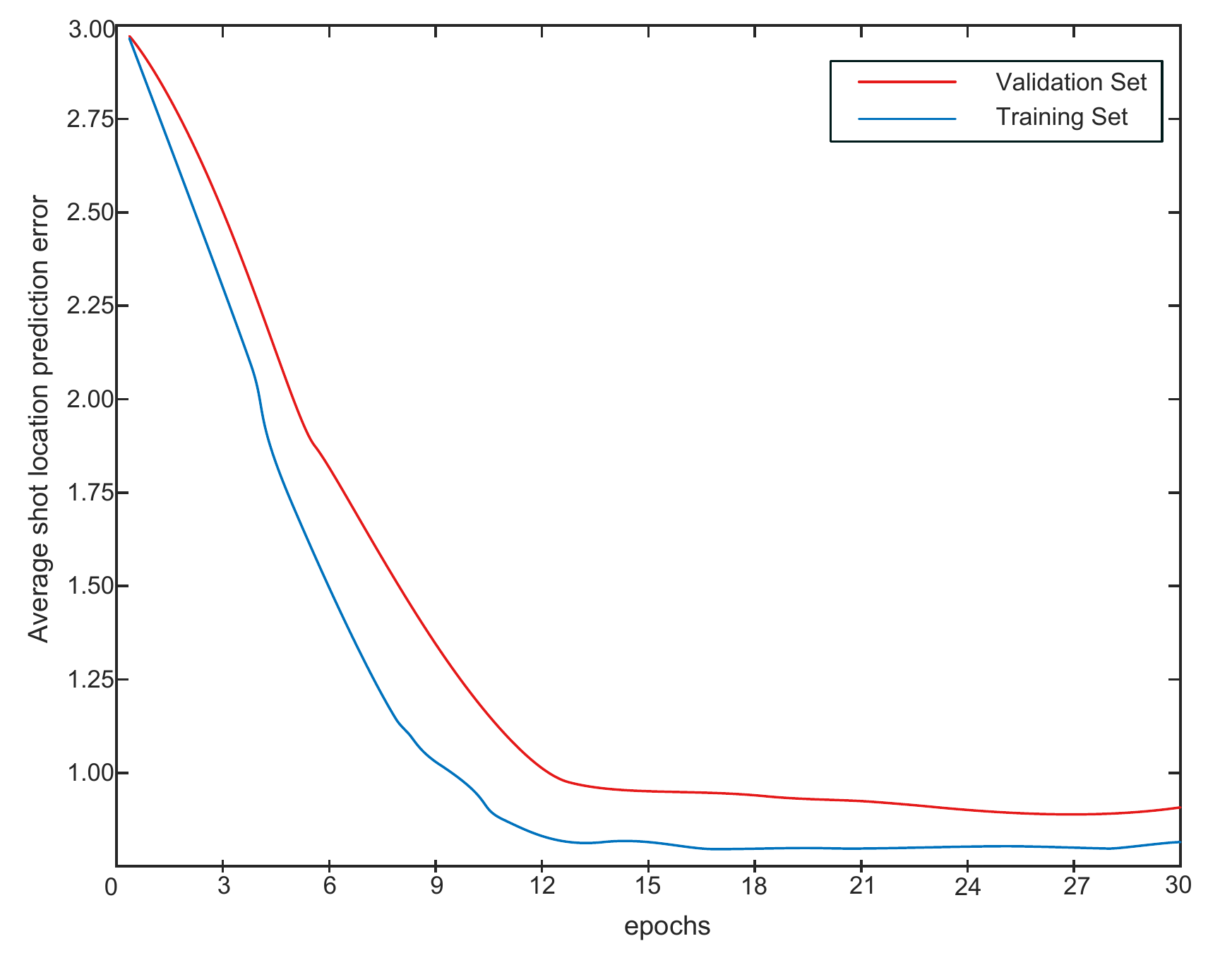}
\caption{Learning curves for training and validation sets for shot location prediction Error}
\label{fig:learning_curves}
\end{figure}

The implementation of the MSS-GAN module presented in this paper is completed using Keras \cite{chollet2017keras} with the Theano \cite{bergstra2010theano} backend. We choose batch size to be 32 and trained the model using the Adam optimiser \cite{kingma2014adam} with a learning rate of 0.002 for 10 epochs and set the learning rate to be 0.0002 for another 20 epochs. Hyper parameters $l$, $N$ and $b$ are evaluated experimentally. Using the validation set of Sec. \ref{sec:shot_location_prediction_sec} we fine tuned each parameter individually, holding the rest of the parameters constant. The experimental evaluations are illustrated in Fig. \ref{fig:hyper_parameters}. As $N=1100$, $l=3$ and $b=80$ gives us the minimum error values, we set the respective parameters accordingly. Fig. \ref{fig:learning_curves} shows learning curves of the proposed MSS-GAN model showing the model convergence. We note that the model doesn't overfit. 

\subsection{Time Efficiency}
The proposed MSS-GAN model doesn't require any special hardware such as GPUs to run and has 33.7M trainable parameters. We ran the test set of Sec. \ref{sec:shot_location_prediction_sec} on a single core of an Intel Xeon E5-2680 2.50GHz CPU and the algorithm was able to generate 1000 shot location predictions 28.5 seconds.

\subsection{Application: Tactics analysis}

\begin{figure}[htbp]
\centering

\subfigure[Djokovic to Federer]{\includegraphics[width = .3 \linewidth]{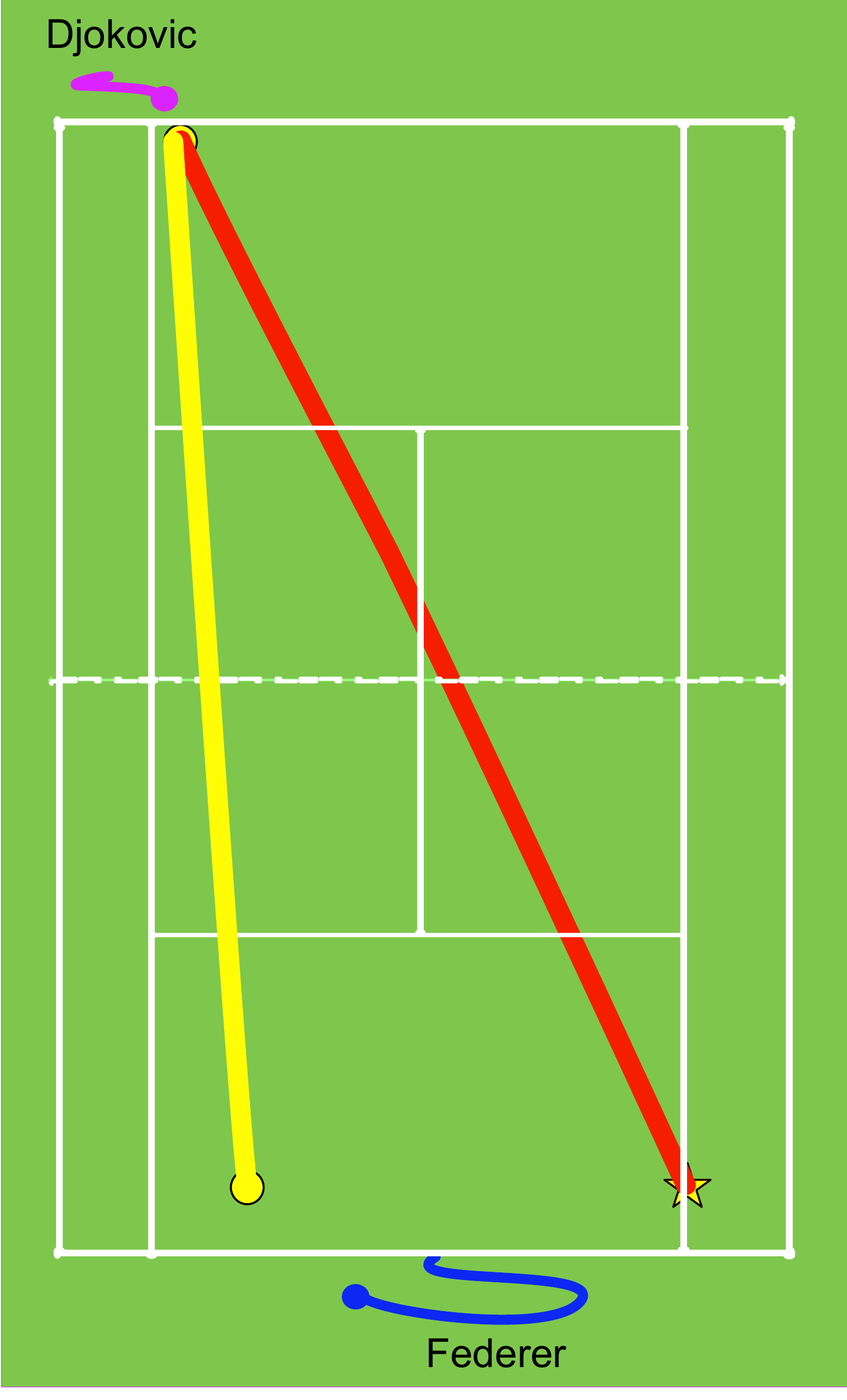}}
\subfigure[Djokovic to Nadal]{\includegraphics[width = .3 \linewidth]{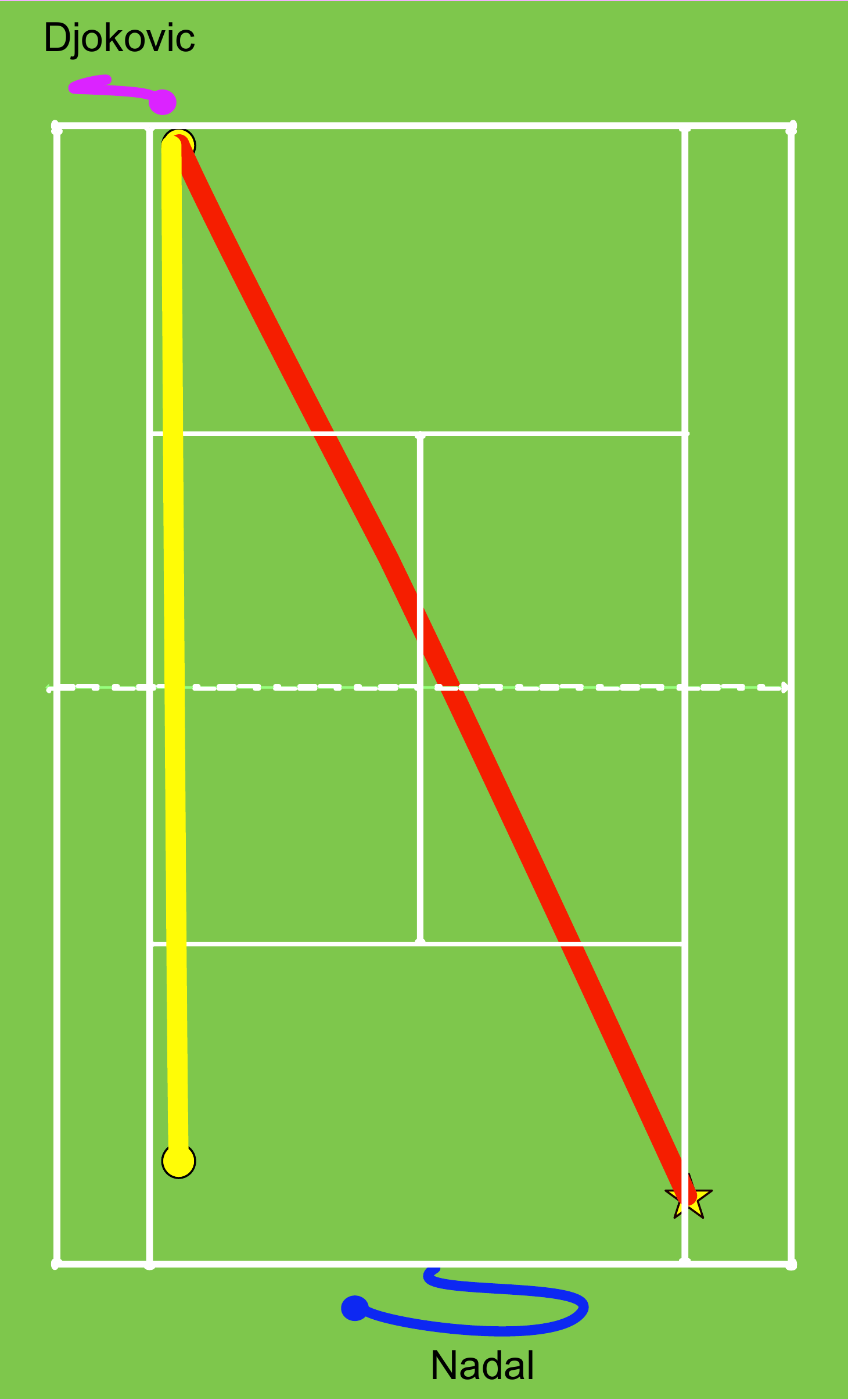}}
\subfigure[Nadal to Federer]{\includegraphics[width = .3 \linewidth]{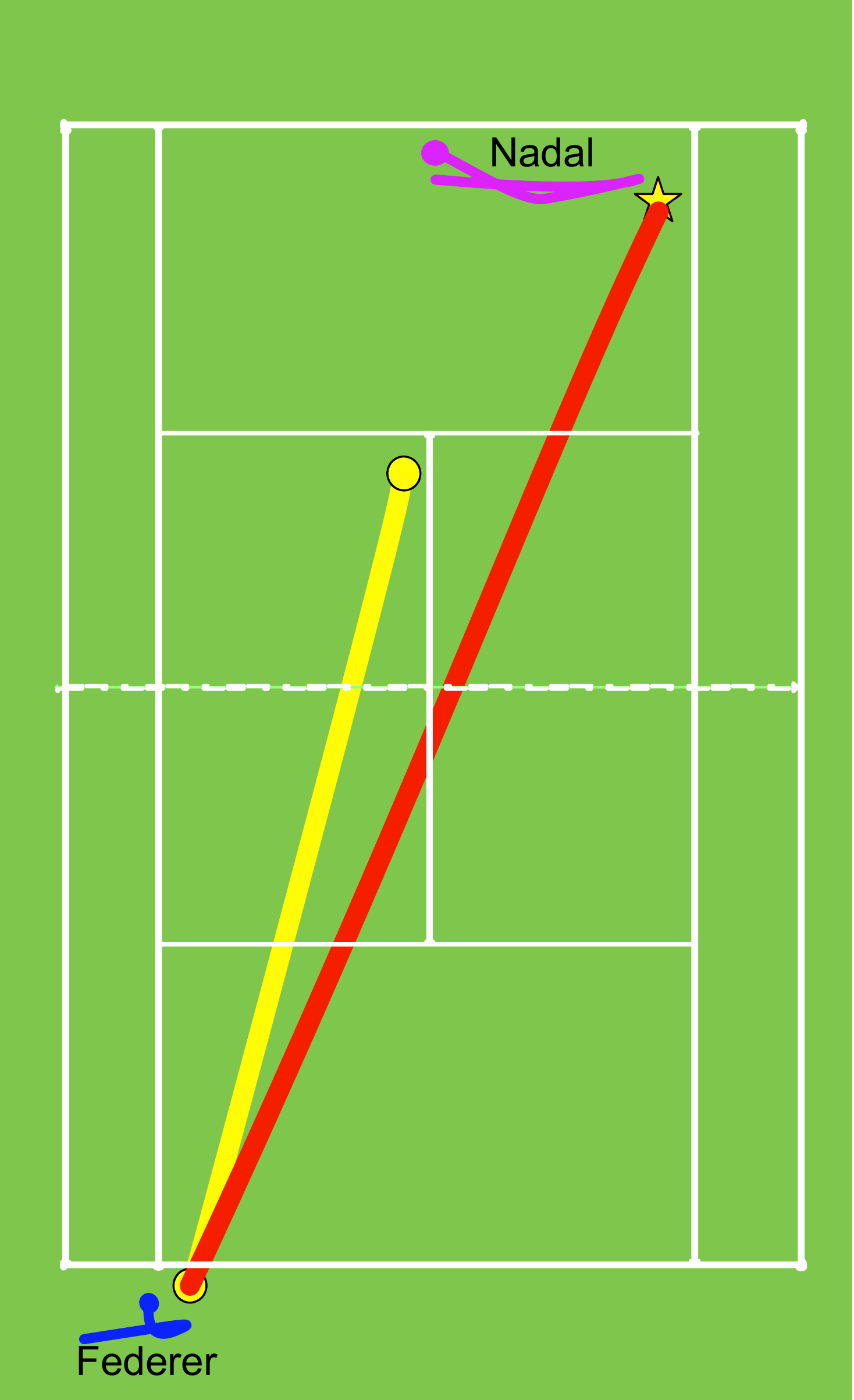}}
\subfigure[Nadal to Djokovic]{\includegraphics[width = .3 \linewidth]{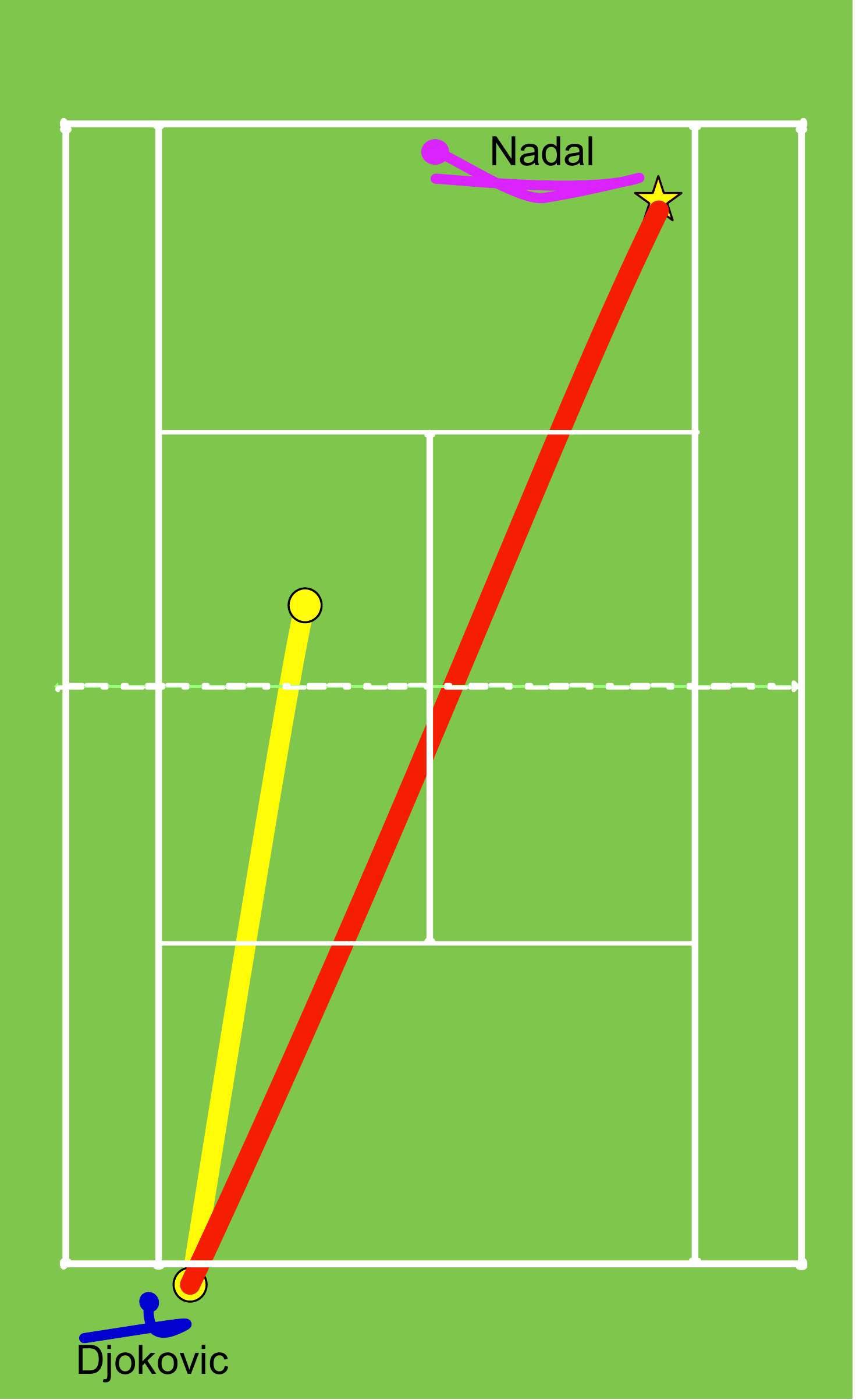}}
\subfigure[Federer to Djokovic]{\includegraphics[width = .3 \linewidth]{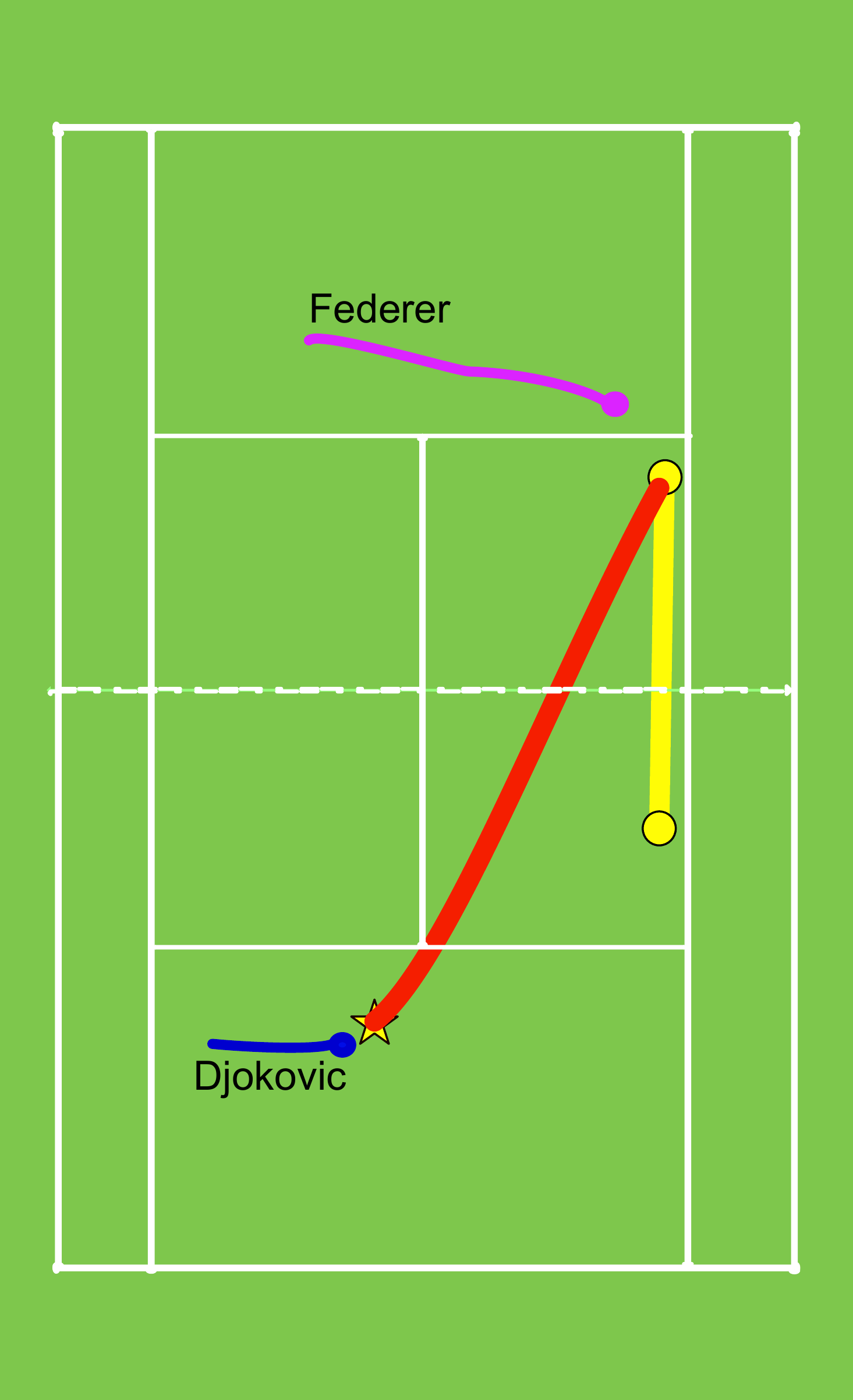}}
\subfigure[Federer to Nadal]{\includegraphics[width = .3 \linewidth]{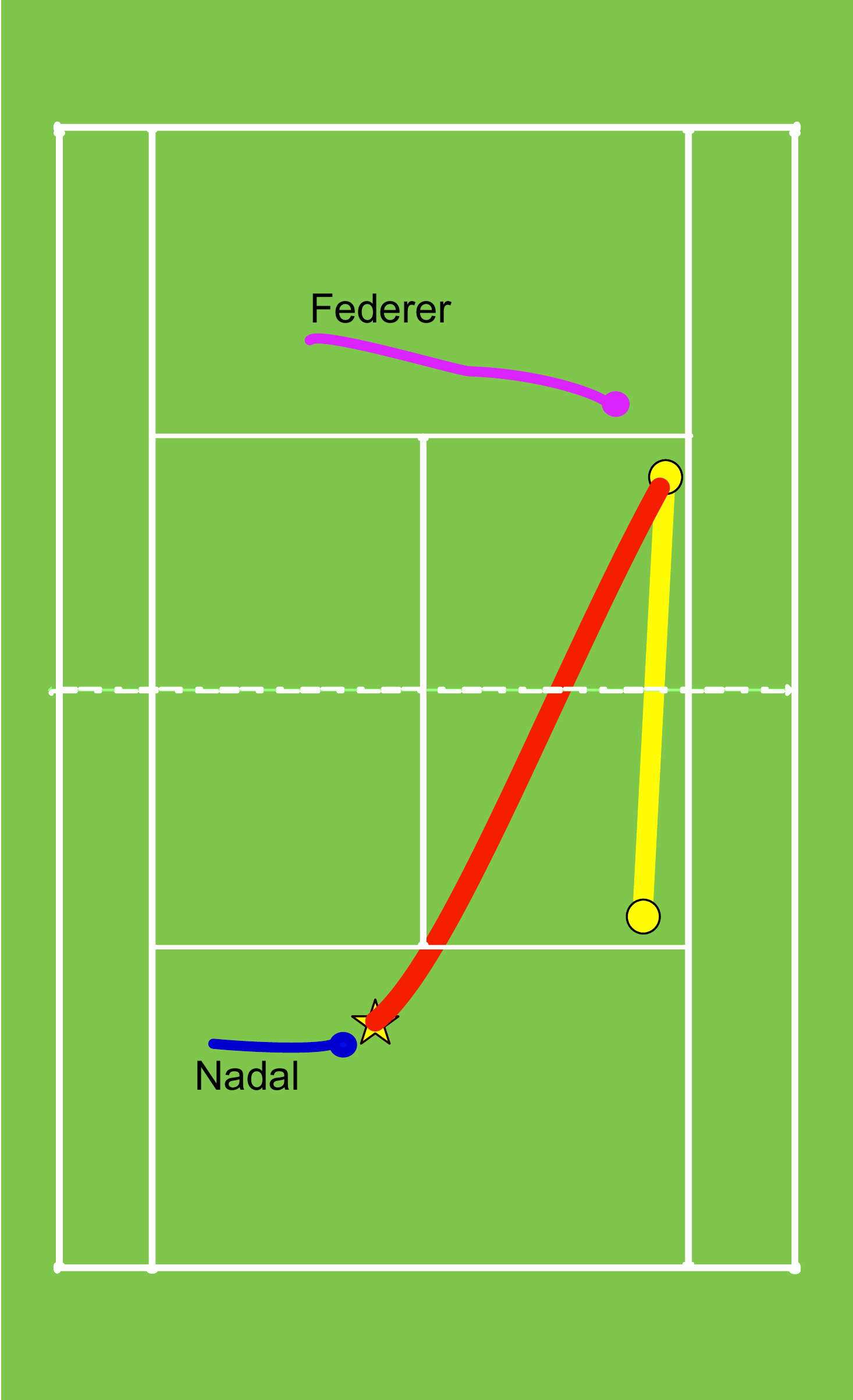}}
\caption{Given the same incoming shot, opponent and player locations, speed ($s_t$), angle ($a_t$) and points ($p_t$), we can change the opponent id ($op_t$) and see how the player of interest changes his strategy to adapt to the opponent. Incoming shot trajectory is denoted in red where the yellow start and circle defines the starting and ending locations. The predicted return shot trajectory is denoted in yellow line where the ending location is represented in a yellow circle. Observed feet movements for the player of interest and opponent are denoted in magenta and blue colours.}
\label{fig:fig_opponent_adaptation}
\end{figure}

The conditional nature of the proposed model allows us to directly infer the opponent adaptation strategies against different players. In Fig \ref{fig:fig_opponent_adaptation} we visualise the predicted return shots for the same incoming shot, and the same context for Djokovic, Nadal and Federer against different opponents. We held the shot location, player and opponent location, incident speed ($s_t$) and angle ($a_t$) and the points ($p_t$) constant and changed only the opponent id $(op_t)$ in the system. This allows us to infer different opponent adaptation strategies employed, and we can explore the way that a given will player will subtlety vary their play style depending on the opponent. From Fig. \ref{fig:fig_opponent_adaptation} we can see variations in the depth and angle of return shots, as players vary the tactics in response to their opponent. 

\begin{figure}[htbp]
\centering
\subfigure[Djokovic to Federer P:00-00]{\includegraphics[width = .3 \linewidth]{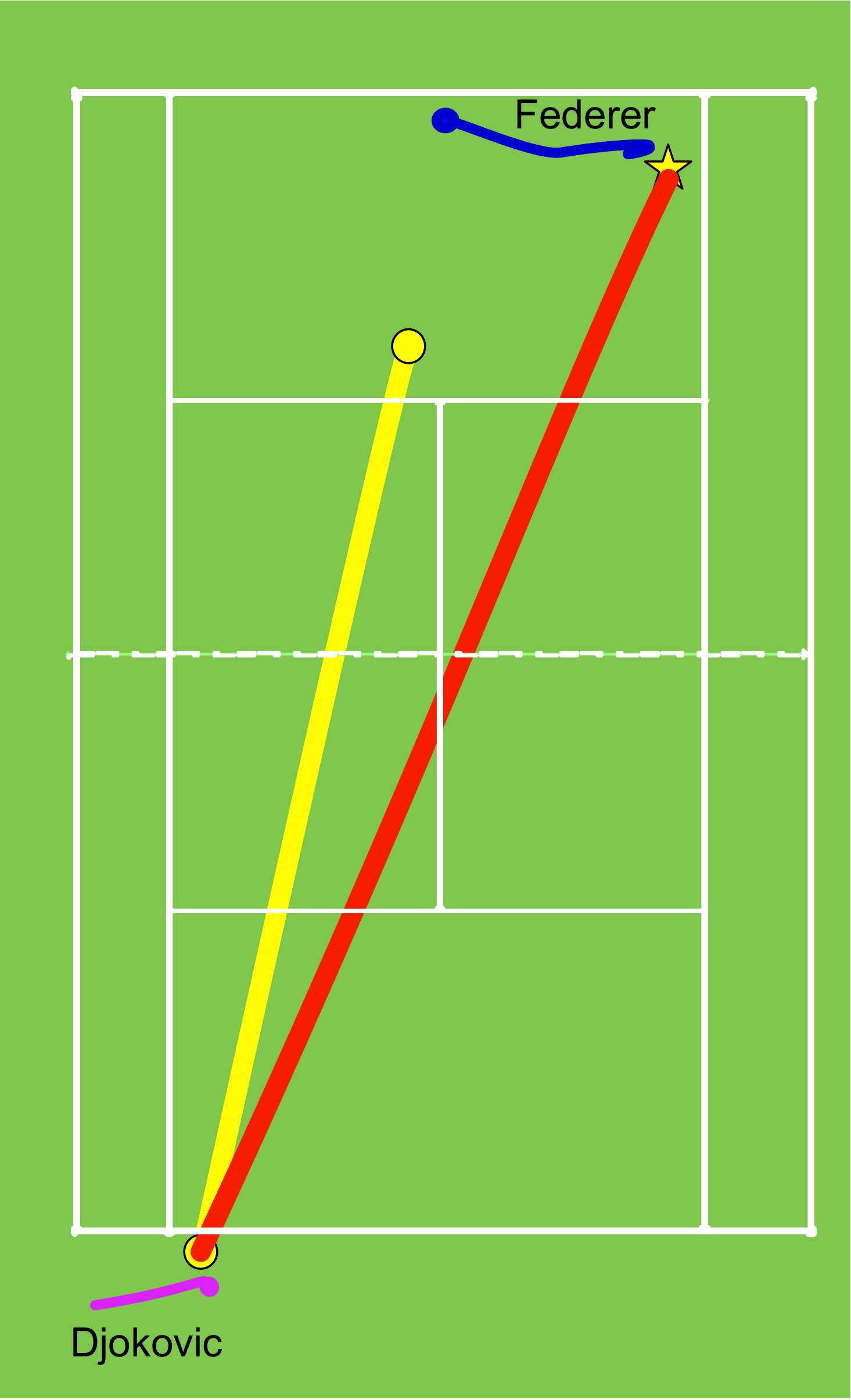}}
\subfigure[Djokovic to Federer P:15-00]{\includegraphics[width = .3 \linewidth]{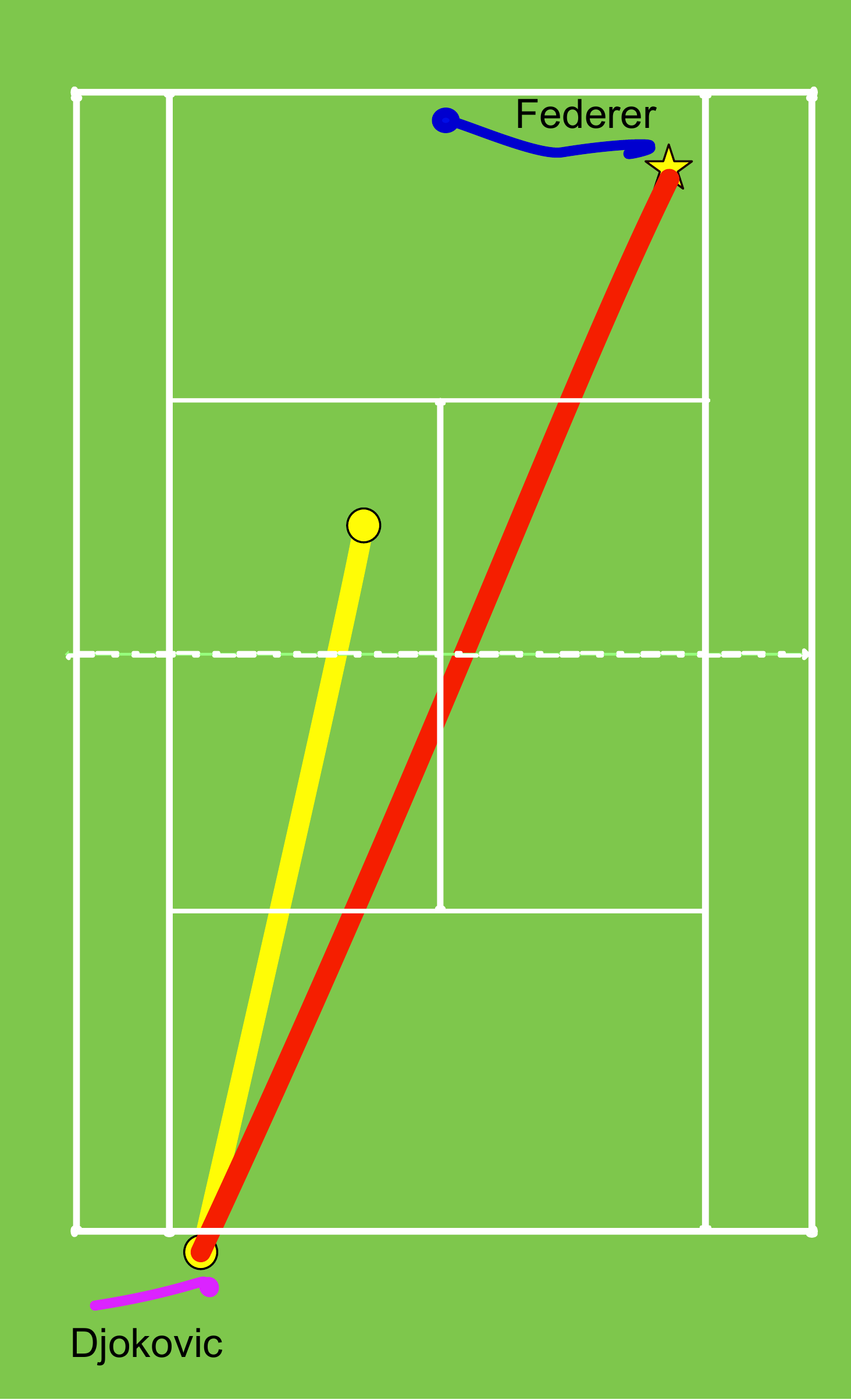}}
\subfigure[Djokovic to Federer P:00-15]{\includegraphics[width = .3 \linewidth]{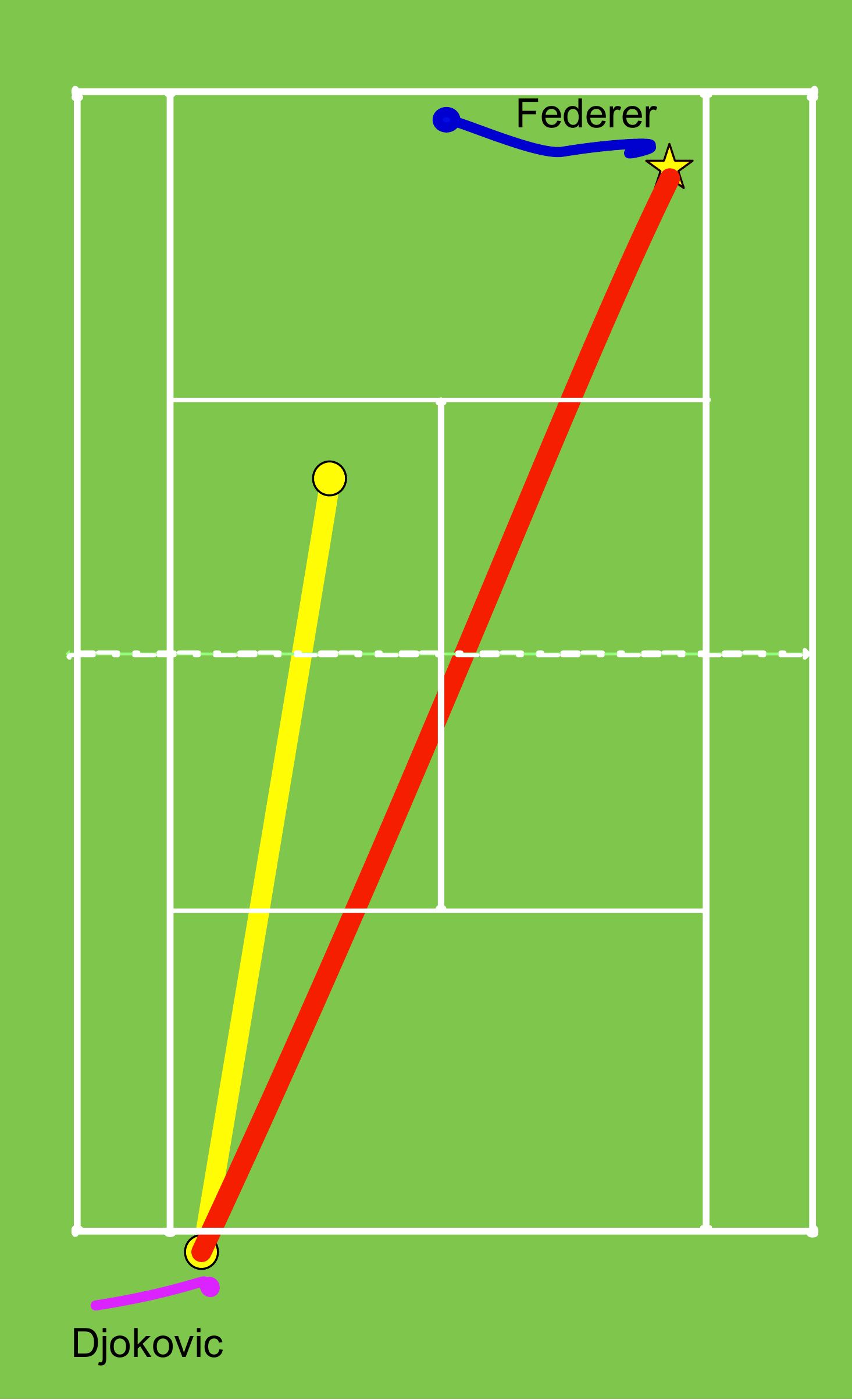}}
\subfigure[Djokovic to Federer P:30-00]{\includegraphics[width = .3 \linewidth]{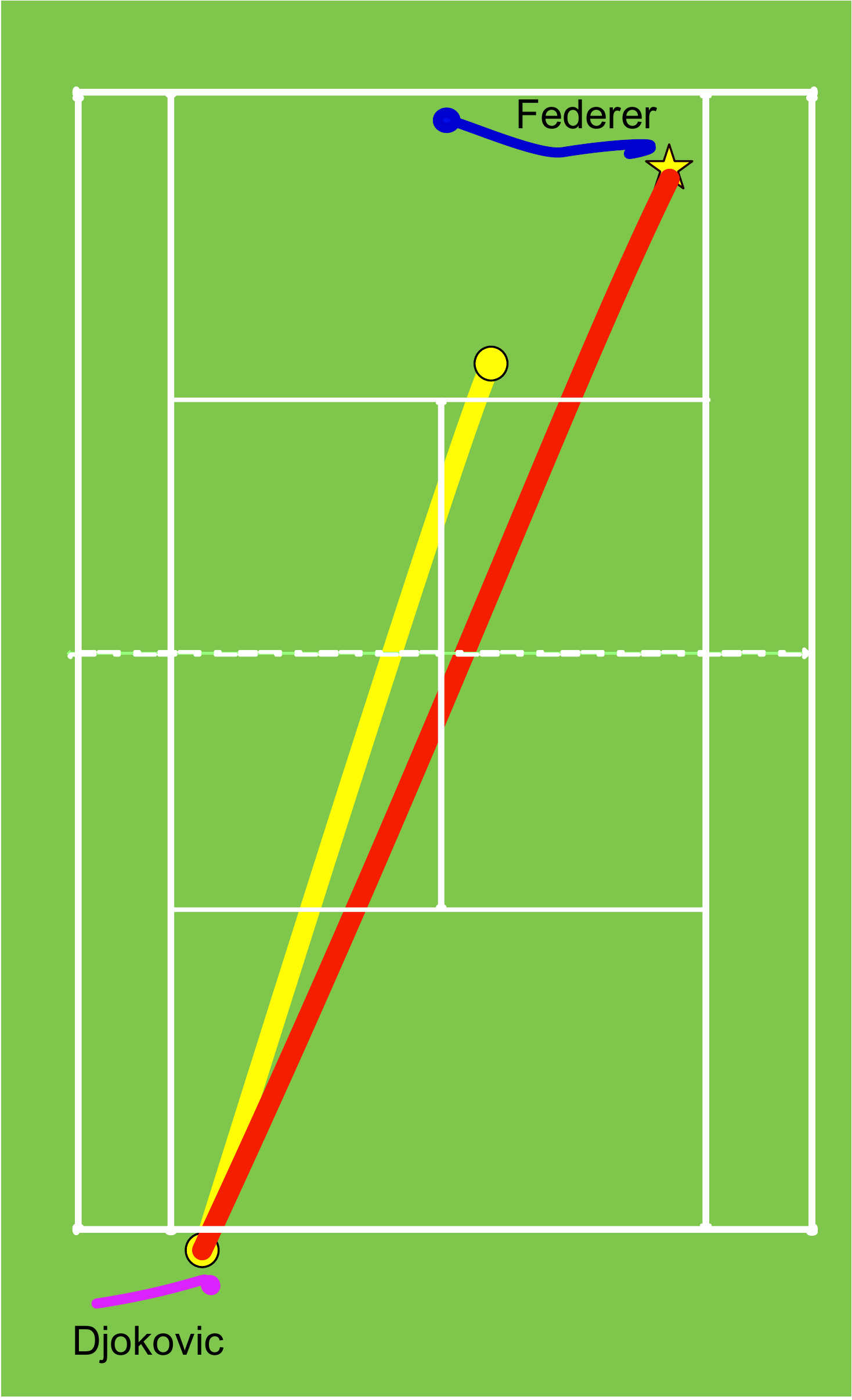}}
\subfigure[Djokovic to Federer P:30-15]{\includegraphics[width = .3 \linewidth]{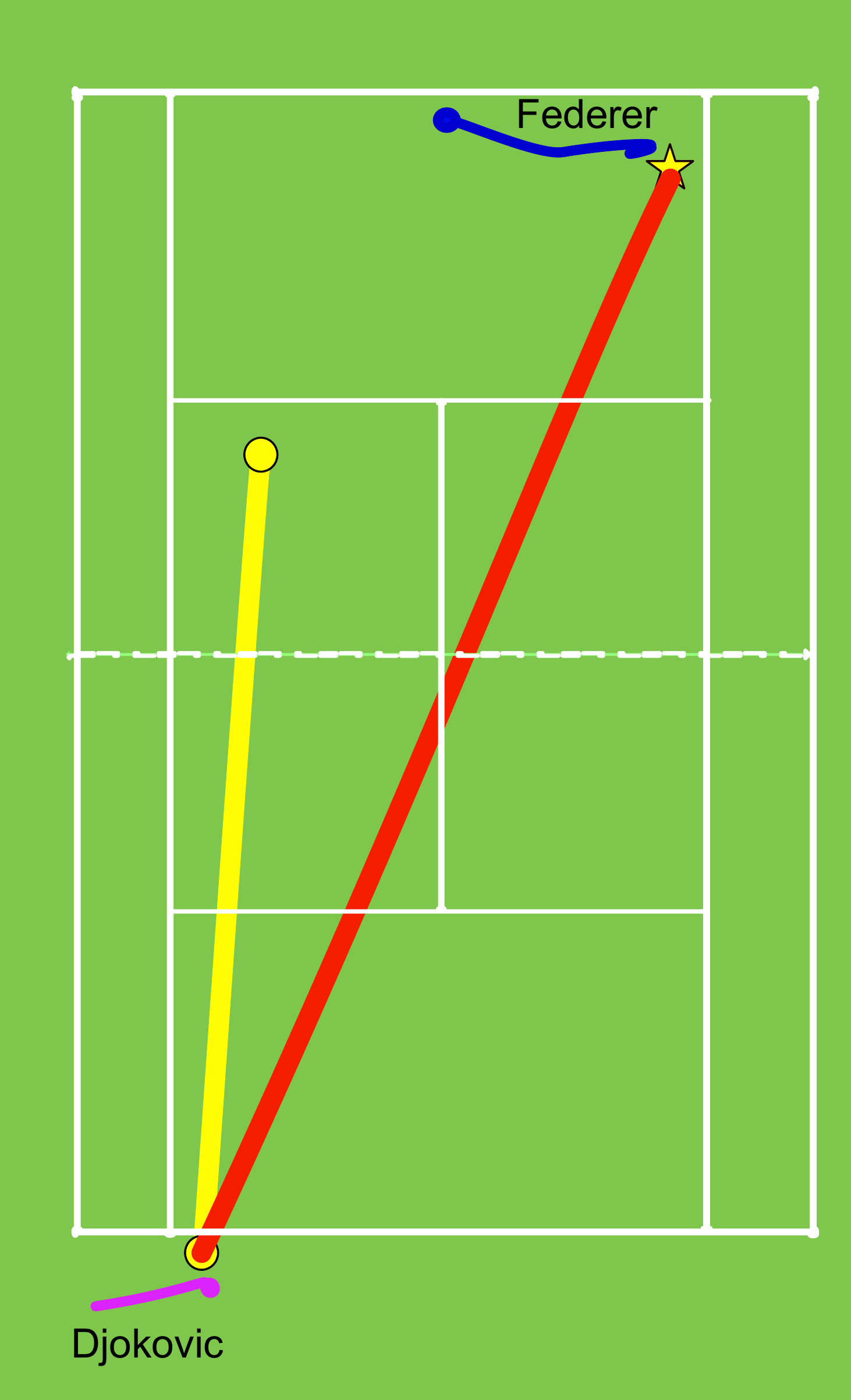}}
\subfigure[Djokovic to Federer P:15-30]{\includegraphics[width = .3 \linewidth]{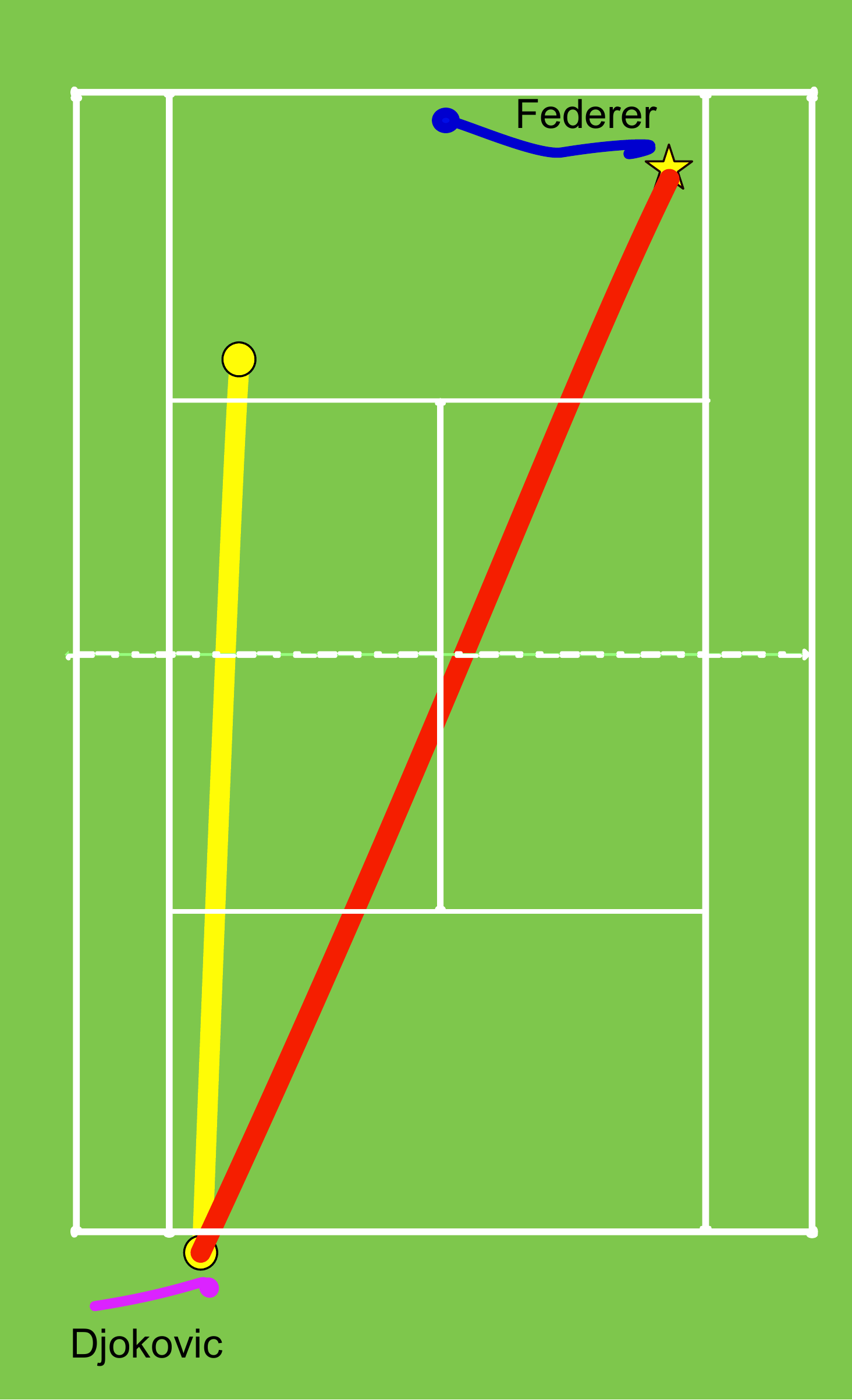}}
\caption{Given the same incoming shot, opponent and player locations, speed ($s_t$), angle ($a_t$) and  opponent id ($op_t$), we can change the points ($p_t$) and see how the player of interest changes his strategy to adapt to the current context. Incoming shot trajectory is denoted in red where the yellow start and circle defines the starting and ending locations. The predicted return shot trajectory is denoted in yellow line where the ending location is represented in a yellow circle. Observed feet movements for the player of interest and opponent are denoted in magenta and blue colours.}
\label{fig:fig_score_adaptation}
\end{figure}

To further demonstrate the importance and value of the proposed context modelling scheme, we investigate how players change their strategies depending on the current score. In Fig. \ref{fig:fig_score_adaptation} we visualise the shot location predictions where we held the shot location, player and opponent location, incident speed ($s_t$) and angle ($a_t$) and the opponent id $(op_t)$ constant and changed only the points ($p_t$) in the system. We see clear differences in the return shot as the score changes, suggesting the has learned when to be aggressive within not only a point, but in the wider context of the match.

These examples demonstrate that the proposed MSS-GAN model is capable of capturing match context and the player tactical elements which are essential when anticipating player behaviour.

\section{Conclusion}
In this paper we propose a method to anticipate the next shot type and location in tennis, by analysing the structure of the player behaviour's and it's temporal accordance. We contribute a novel data driven method to capture salient information from the observed game context and propose methodologies to capture longterm historical experiences of different players, emulating the episodic memory behaviour of the human brain. Additionally, we introduce a novel methodology for learning abstract level concepts through a tree structured episodic memory and propose methods for transferring this acquired knowledge to a neural semantic memory component. Our quantitative and qualitative evaluations on a tennis player tracking dataset from the 2012 Australian Men's Open demonstrate the capacity of the proposed method to anticipate complex real world player strategies, and its potential to be applied to other data mining/ knowledge discovery applications.

%

\ifCLASSOPTIONcompsoc
  \section*{Acknowledgments}
\else
  \section*{Acknowledgment}
\fi

The authors would like to thank Tennis Australia for providing access to the Hawk-eye player tracking database for this analysis. The authors also thank QUT High Performance Computing (HPC) for providing the computational resources for this research.

\ifCLASSOPTIONcaptionsoff
  \newpage
\fi



\bibliographystyle{IEEEtran}
\bibliography{egbib.bib}

\begin{IEEEbiography}[{\includegraphics[width=1in,height=1.25in,clip,keepaspectratio]{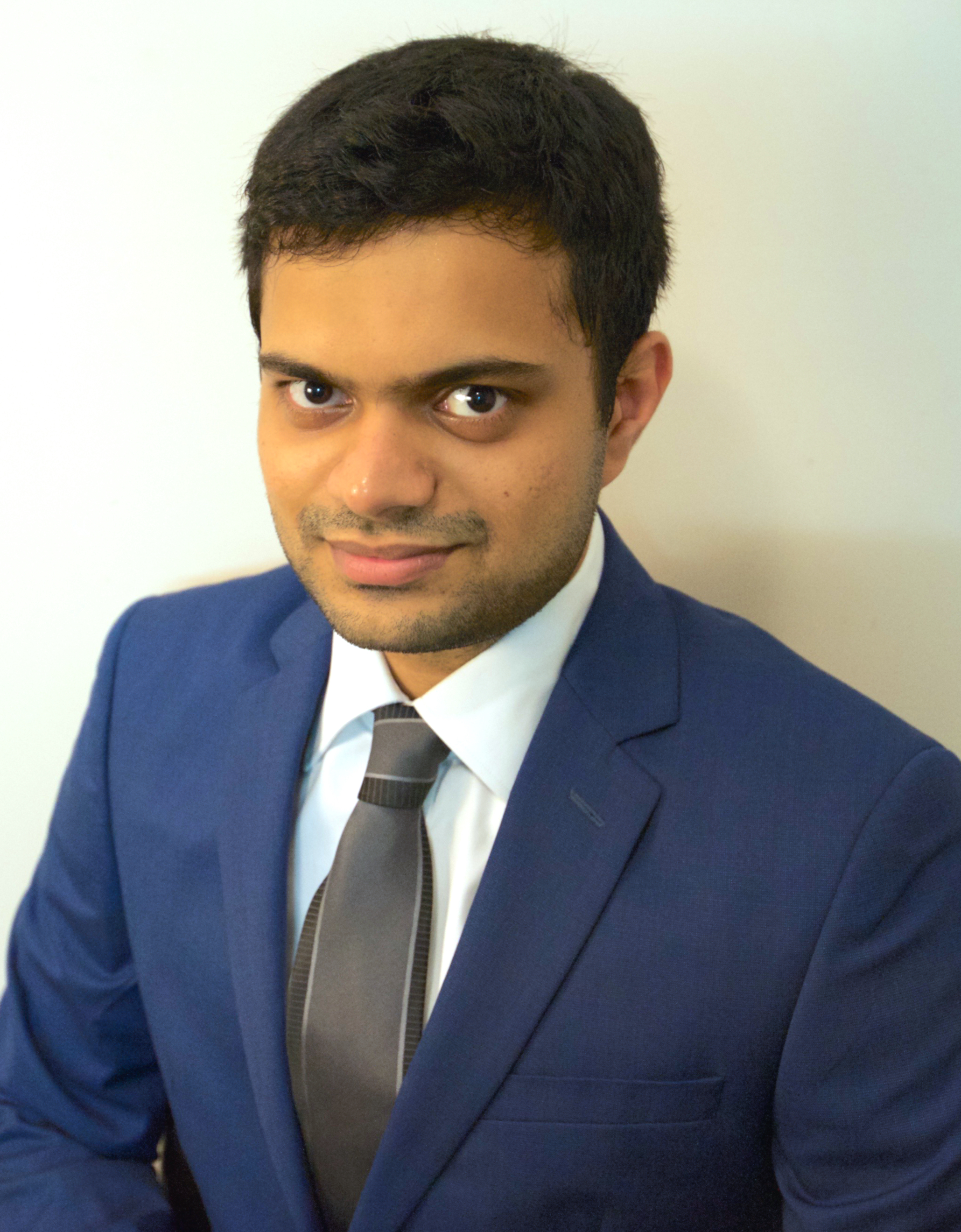}}]{Tharindu Fernando} is a PhD student at Queensland University of Technology, Australia. He received his Bachelor of Computer Science with first class honours from University of Peradeniya, Sri Lanka, in 2015. Prior to the beginning of his PhD program, he has conducted variety of research projects, resulting in automated systems to evaluate player biomechanics and perform strategic analysis in sports. His research interests focus mainly onto human behaviour analysis and prediction. 
\end{IEEEbiography}

\begin{IEEEbiography}[{\includegraphics[width=1in,height=1.25in,clip,keepaspectratio]{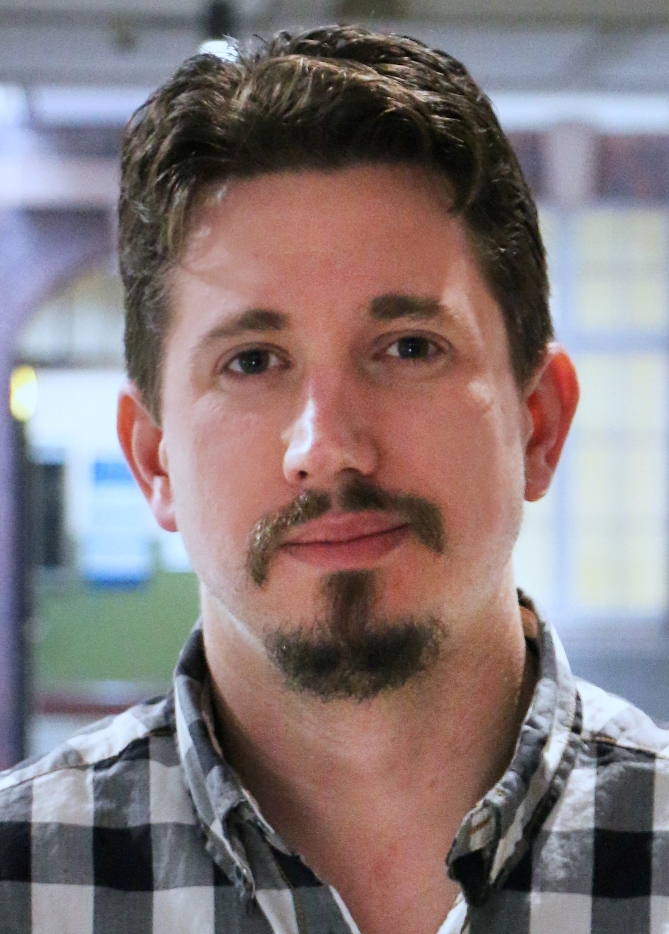}}]{Dr. Simon Denman} received a BEng (Electrical), BIT, and PhD in the area of object tracking from the Queensland University of Technology (QUT) in Brisbane, Australia. He is currently a Senior Research Fellow with the Speech, Audio, Image and Video Technology Laboratory at QUT. His active areas of research include intelligent surveillance, video analytics, and video-based recognition.
\end{IEEEbiography}

\begin{IEEEbiography}[{\includegraphics[width=1in,height=1.25in,clip,keepaspectratio]{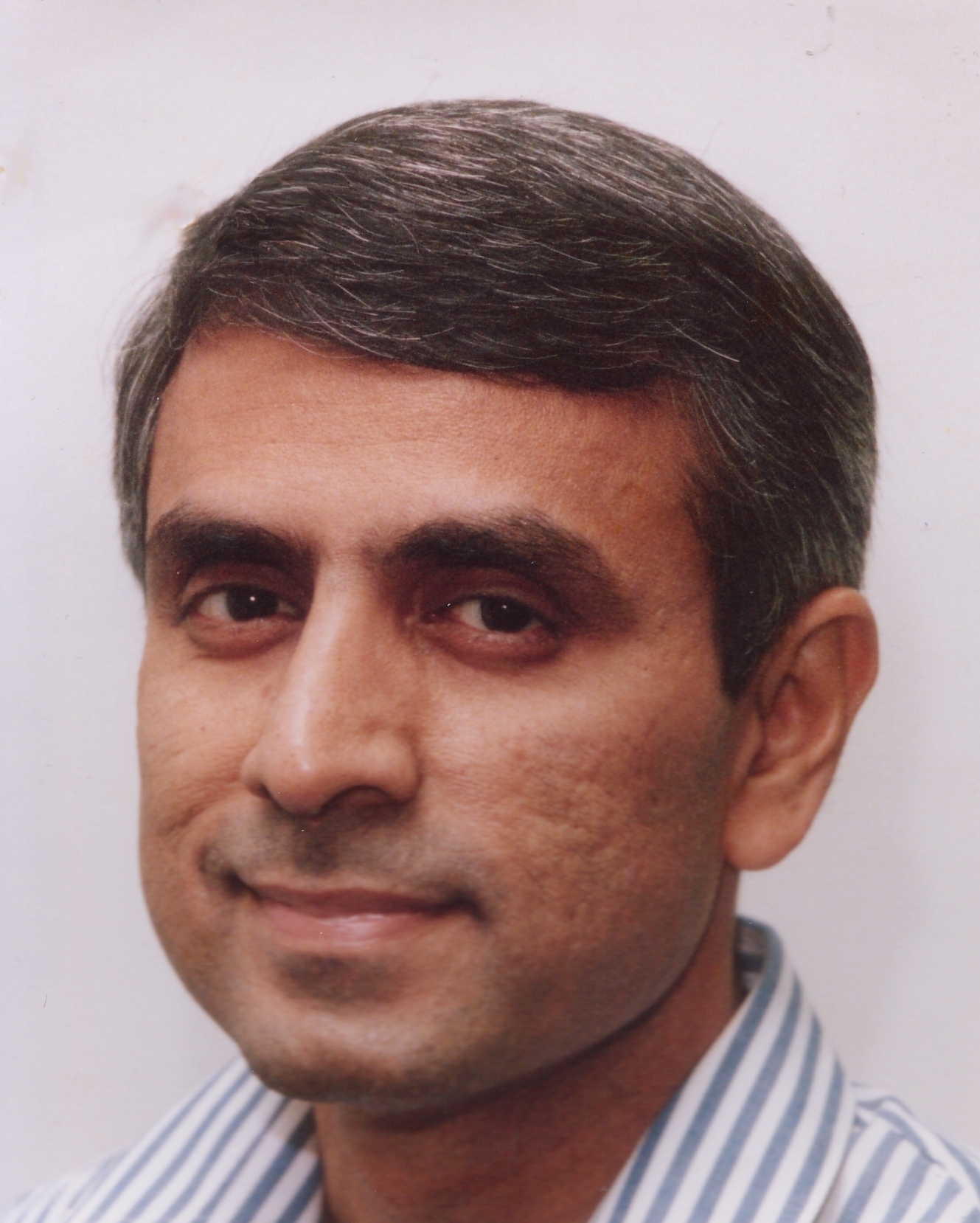}}]{Professor Sridha Sridharan} has a BSc (Electrical Engineering) degree and obtained a MSc (Communication Engineering) degree from the University of Manchester, UK and a PhD degree from University of New South Wales, Australia. He is currently with the Queensland University of Technology (QUT) where he is a Professor in the School Electrical Engineering and Computer Science. Professor Sridharan is the Leader of the Research Program in Speech, Audio, Image and Video Technologies (SAIVT) at QUT, with strong focus in the areas of computer vision, pattern recognition and machine learning. He has published over 500 papers consisting of publications in journals and in refereed international conferences in the areas of Image and Speech technologies during the period 1990-2016. During this period he has also graduated 60 PhD students in the areas of Image and Speech technologies. Prof Sridharan has also received a number of research grants from various funding bodies including Commonwealth competitive funding schemes such as the Australian Research Council (ARC) and the National Security Science and Technology (NSST) unit. Several of his research outcomes have been commercialised.
\end{IEEEbiography}

\begin{IEEEbiography}[{\includegraphics[width=1in,height=1.25in,clip,keepaspectratio]{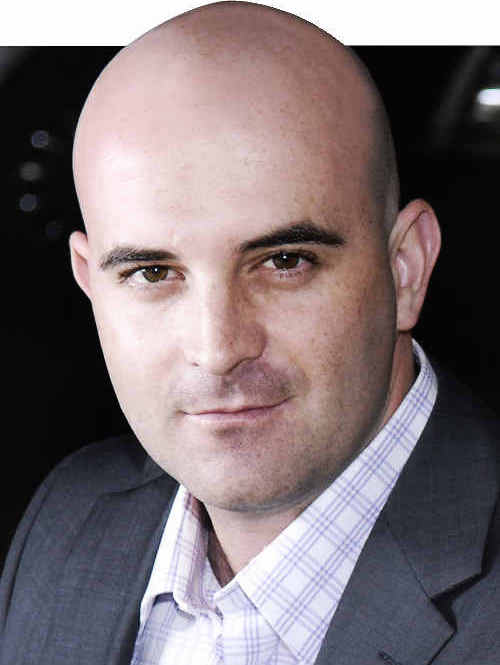}}]{Clinton Fookes} is a Professor in Vision Signal Processing and the Speech, Audio, Image and Video Technologies group within the Science and Engineering Faculty at QUT. He holds a BEng (Aerospace/Avionics), an MBA with a focus on technology innovation/management, and a PhD in the field of computer vision. Clinton actively researches in the fields of computer vision and pattern recognition including video surveillance, biometrics, human- computer interaction, airport security and operations, command and control, and complex systems. Clinton has attracted over $\$15M$ of cash funding for fundamental and applied research from external competitive sources and has published over 140 internationally peer- reviewed articles. He has been the Director of Research for the School of Electrical Engineering and Computer Science. He is currently the Head of Discipline for Vision Signal Processing. He is the Technical Director for the Airports of the Future collaborative research initiatives. He is a Senior Member of the IEEE, and a member of other professional organisations including the APRS.
\end{IEEEbiography}

\end{document}